\documentclass[10pt,twocolumn,letterpaper]{article}

\usepackage[accsupp]{axessibility}
\usepackage[pagenumbers]{cvpr} 
\usepackage{url}            
\usepackage{booktabs}       
\usepackage{amsfonts}       
\usepackage{nicefrac}       
\usepackage{microtype}      
\usepackage{amsmath, amssymb, amsthm}
\usepackage{colortbl}
\usepackage{pifont}
\usepackage{multirow}
\usepackage{multicol}
\usepackage{makecell}
\usepackage{graphicx}
\usepackage{diagbox}
\usepackage{subcaption}
\usepackage[utf8]{inputenc}
\usepackage{svg}
\usepackage{comment} 

\usepackage{float}
\usepackage{enumitem}
\usepackage{caption}
\usepackage{graphicx}
\usepackage{caption}
\usepackage{booktabs}
\usepackage{graphicx}
\usepackage{caption}
\usepackage{subcaption}
\usepackage{newunicodechar}
\newunicodechar{，}{,}
\definecolor{cvprblue}{rgb}{0.21,0.49,0.74}
\usepackage[pagebackref,breaklinks,colorlinks,allcolors=cvprblue]{hyperref}

\def\paperID{37968} 
\def\confName{CVPR}
\def\confYear{2026}

\title{LESA: Learnable Stage-Aware Predictors for Diffusion Model Acceleration}

\def\lessspaces{~~~}
\author{
Peiliang Cai\footnotemark[1], \lessspaces{}
Jiacheng Liu\footnotemark[1], \lessspaces{}
Haowen Xu, \lessspaces{}
Xinyu Wang, \lessspaces{}
Chang Zou, \lessspaces{}
Linfeng Zhang \footnotemark[2] \\\\
Shanghai Jiao Tong University
}

\begin{document}
\raggedbottom

\twocolumn[{%
\maketitle
\begin{figure}[H]
\vspace{-9mm}
\hsize=\textwidth
\centering
\includegraphics[width=2\linewidth]{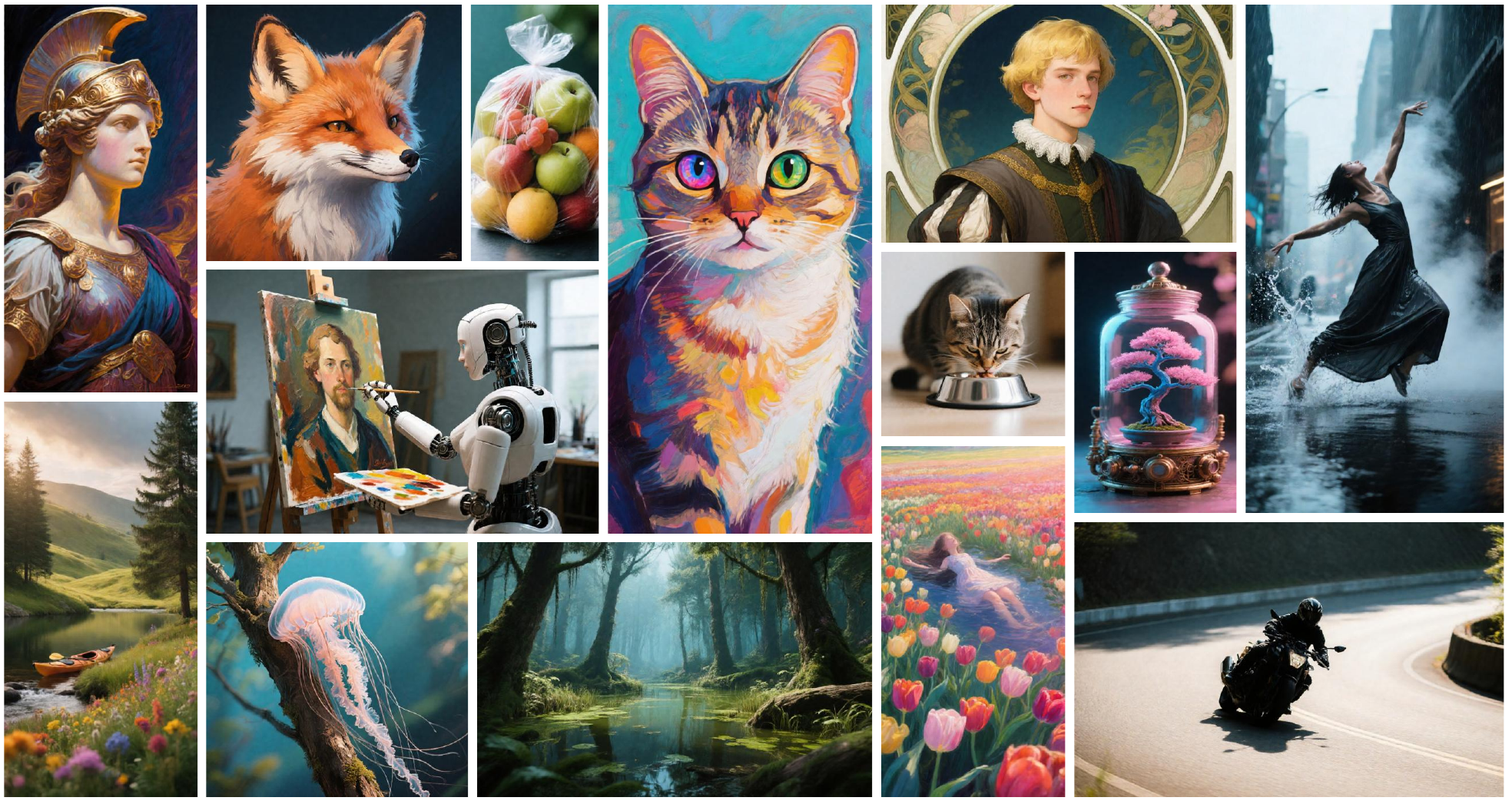}
\vspace{-2mm}

    \caption{Images generated by Qwen-image using \textit{LESA} with a 6.25$\times$ acceleration.}
    \label{fig:showcase}

\end{figure}
}]
\renewcommand{\thefootnote}{\fnsymbol{footnote}}
\footnotetext[1]{Equal contribution.}
\footnotetext[2]{Corresponding author.}

\begin{abstract}
Diffusion models have achieved remarkable success in image and video generation tasks. However, the high computational demands of Diffusion Transformers (DiTs) pose a significant challenge to their practical deployment. While feature caching is a promising acceleration strategy, existing methods based on simple reusing or training-free forecasting struggle to adapt to the complex, stage-dependent dynamics of the diffusion process, often resulting in quality degradation and failing to maintain consistency with the standard denoising process. To address this, we propose a \textbf{LE}arnable \textbf{S}tage-\textbf{A}ware (\textbf{LESA}) predictor framework based on two-stage training. Our approach leverages a Kolmogorov–Arnold Network (KAN) to accurately learn temporal feature mappings from data. We further introduce a multi-stage, multi-expert architecture that assigns specialized predictors to different noise-level stages, enabling more precise and robust feature forecasting. Extensive experiments show our method achieves significant acceleration while maintaining high-fidelity generation. Experiments demonstrate 5.00$\times$ acceleration on FLUX.1-dev with minimal quality degradation (1.0\% drop), 6.25$\times$ speedup on Qwen-Image with a 20.2\% quality improvement over the previous SOTA (TaylorSeer), and 5.00$\times$ acceleration on HunyuanVideo with a 24.7\% PSNR improvement over TaylorSeer. State-of-the-art performance on both text-to-image and text-to-video synthesis validates the effectiveness and generalization capability of our training-based framework across different models. Our code is available at \url{https://github.com/caipeiliang2004/LESA}.
\end{abstract}    
\vspace{-10mm}
\section{Introduction}
\vspace{-1mm}
\label{sec:intro}

\begin{figure*}[htbp]
    \centering
    \includegraphics[width=\linewidth]
    {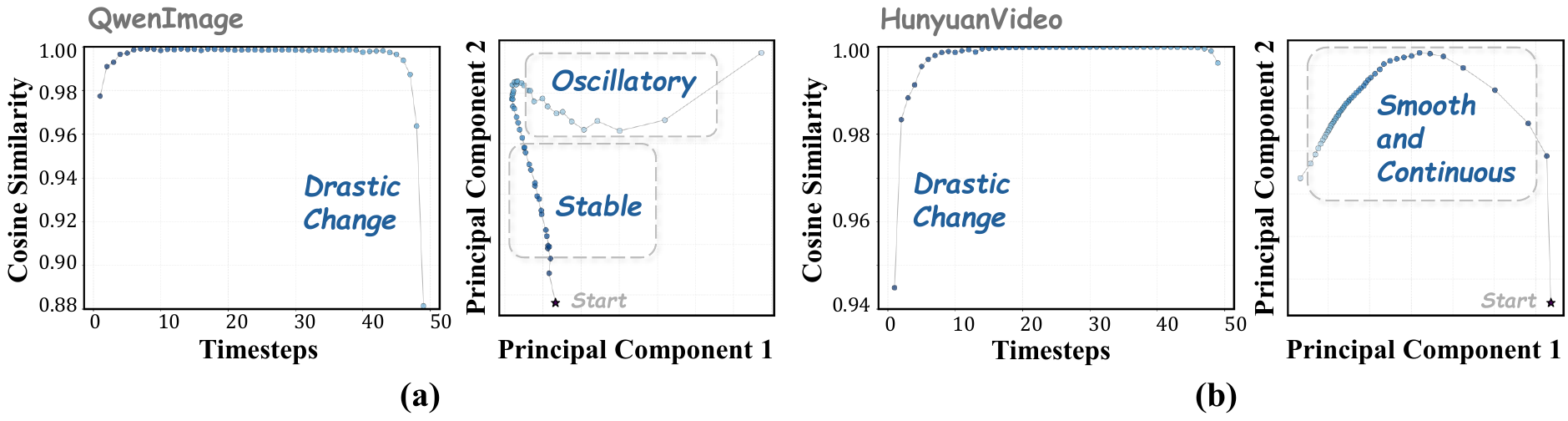}
    \vspace{-7mm}
\caption{\textbf{Cosine Similarity curves and PCA-projected trajectories} show that feature evolution differs markedly across diffusion models, with non-smooth, stage-dependent dynamics that challenge the common assumption of simple or continuous temporal change.}

    \label{fig:cosine_similarity_and_pca}
    \vspace{-5mm}
\end{figure*}

Diffusion Models (DMs)\citep{DM,StableDiffusion,blattmann2023SVD} have achieved remarkable success in image and video generation tasks. Recently, Diffusion Transformers (DiTs) \citep{DiT} have further improved the quality and diversity of generated content. However, DiT models face significant computational challenges during inference due to their deep transformer architectures and the need for iterative denoising across multiple timesteps. These limitations restrict their practical deployment in real-world applications. To address this efficiency problem, researchers have proposed a new paradigm known as Feature Caching, which exploits temporal redundancy between consecutive denoising steps to accelerate model inference~\citep{ma2024deepcache,li2023FasterDiffusion,chen2024delta-dit, liu2025survey}.

Current feature caching methods can be divided into two categories: \textbf{``Cache-Then-Reuse''} and \textbf{``Cache-Then-Forecast''}. The former relies on the assumption that feature variations between adjacent timesteps in diffusion models are minimal, and directly reuses the features from the previous timestep~\citep{selvaraju2024fora,zou2024accelerating}. In contrast, the latter assumes that \textit{feature evolution of the diffusion process is smooth over time} and uses\emph{training-free} techniques, such as Taylor expansion, to extrapolate features for future timesteps~\citep{liuTaylorSeer2025}. Although forecasting methods have achieved significant acceleration, their core assumption that features evolve smoothly over time does not always hold in the actual diffusion process.

In our analysis, we observe that the feature dynamics of diffusion models exhibit a \textbf{\emph{stage-dependent}} property as shown in Fig.~\ref{fig:cosine_similarity_and_pca}. The diffusion process starts with a high-noise stage characterized by drastic and unstable feature changes, and ends with a detail-refinement stage where fine-grained features are progressively reconstructed. In contrast, the middle stage shows more stable and continuous feature evolution. Consequently, conventional caching strategies that adopt  \textbf{\textit{a single fixed prediction scheme}} across all timesteps are inherently \emph{incapable of adapting to the stage-varying dynamics of the diffusion process}, leading to degraded generation quality and temporal inconsistency.

To address these issues, several time-aware caching strategies have been proposed. Methods like TeaCache~\cite{liu2024timestep} and SpeCa~\cite{Liu2025SpeCa} adaptively adjust the caching interval to better accommodate the non-uniform dynamics of the diffusion process. Despite these methods enhancing the flexibility of the caching mechanism to some extent, their core still relies on \emph{heuristic rules} and lacks the capacity to learn the underlying temporal dynamics.


To this end, we propose a \textbf{\emph{training-based stage-aware learnable predictor framework}} that explicitly models temporal characteristics of the diffusion process for efficient and stable inference acceleration. At its core, we introduce a Kolmogorov–Arnold Network (KAN)~\citep{Liu2024KANKN} as the foundational architecture for the learnable predictor, combining strong learning capacity with mathematical elegance and remarkable parameter efficiency.

Furthermore, we propose \textbf{\emph{a multi-stage, multi-expert prediction mechanism}}. This framework partitions the diffusion process into distinct stages based on noise levels, deploying a dedicated expert predictor for each. Specifically, a high-noise expert handles the initial stage of drastic feature change; a mid-noise expert manages the stable, continuous denoising in the middle phase; and a low-noise expert is solely responsible for final detail refinement. This architecture explicitly models the stage-dependent nature of the process, ensuring high-quality generation and temporal consistency even at high acceleration ratios.

Correspondingly, the training procedure also follows a staged strategy. It begins with \textbf{\emph{ground-truth guided training (GT-Guided Training)}} to learn denoising dynamics, and then progresses to \textbf{\emph{closed-Loop autoregressive training (CL-AR Training)}} to build robustness against the inevitable prediction drift encountered during fast, multi-step inference.

\textbf{Our main contributions are as follows:}
\begin{itemize}
\item \textbf{Learnable Temporal Prediction Framework:} We introduce a parameter-efficient predictor based on a Kolmogorov-Arnold Network that directly learns temporal dynamics from diffusion model activations, overcoming the limitations of training-free forecasting methods.
\item \textbf{Stage-Aware Multi-Expert Architecture:} Our framework partitions the denoising process into distinct stages, each handled by a dedicated expert predictor. This specialized design ensures accurate modeling of the unique feature dynamics at different noise levels, enhancing both prediction fidelity and generalization.
\item \textbf{Two-Stage Training Strategy:} We propose a two-stage training methodology that begins with ground-truth guided training to learn denoising dynamics and progresses to closed-loop autoregressive training to build robustness against prediction drift, ensuring stable performance during accelerated inference.
\end{itemize}

\section{Related Works}
\label{sec:related_works}
\vspace{-1mm}
Diffusion models have shown remarkable generative performance across various tasks, like image and video synthesis. As research progresses, their architectures have evolved from convolutional U-Nets~\cite{ronneberger2015unet} to diffusion transformers~\cite{DiT}, shifting from local feature modeling to global context modeling and achieving stronger representation and scalability. However, the improved generation quality of large-scale diffusion models~\cite{yang2025cogvideox,li_hunyuan-dit_2024,kong2024hunyuanvideo,wan_wan_2025} comes at the cost of heavy computation: each denoising process requires dozens of complex network evaluations, leading to high latency and demanding hardware resources. To address this bottleneck, researchers have focused on two main directions: \textit{reducing temporal complexity} and \textit{improving per-step efficiency}.

\subsection{Reducing Temporal Iteration}
Temporal reduction can be approached from two perspectives: \textit{sampling schedulers} and \textit{trajectory compression}.

\vspace{1mm}
\noindent \textbf{Sampling-based Schedulers.} Sampling-based acceleration reduces inference steps while preserving image quality. DDIM~\cite{songDDIM} introduced deterministic sampling with fewer iterations, and the DPM-Solver~\cite{lu2022dpm, lu2022dpm++} series further improved efficiency using high-order ODE solvers. Flow Matching~\cite{refitiedflow} later generalized diffusion into a deterministic transformation between noise and data, enabling faster and more stable sampling. However, their performance is highly dependent on model architecture and noise schedule, making generalization across different frameworks difficult. 

\vspace{1mm}
\noindent \textbf{Trajectory-based Compression.} Instead of repeatedly simulating the entire diffusion trajectory, this line of work trains models to approximate the multi-step denoising behavior in a limited number of steps. Progressive compression~\cite{yan2024perflow} techniques enable few-step inference, while Consistency Models~\cite{song2023consistency} achieve nearly one-step generation by enforcing trajectory consistency across timesteps. However, these methods require additional training, and their acceleration gains are relatively limited compared with other approaches.

\subsection{Enhancing Efficiency in Denosing-Step}
Beyond temporal reduction, another direction aims to enhance per-step efficiency by minimizing redundant computation and memory usage while preserving generation quality.

\vspace{1mm}
\noindent \textbf{Model Simplification and Token Reduction.}Considering the continuous scaling up of model size, some works~\cite{structural_pruning_diffusion, zhu2024dipgo, huang2025diffusion} aim to simplify the model's architecture to reduce computational cost in each inference. Another line of research focuses on reducing the number of tokens fed into the diffusion models, thereby accelerating attention computation. Token pruning~\cite{zhang2024tokenpruningcachingbetter} and adaptive token selection~\cite{saghatchian2025cached} methods dynamically discard visually or semantically redundant tokens while retaining essential contextual information for generation. However, aggressive compression easily breaks feature consistency and harms visual fidelity, making it difficult to accelerate without degrading output quality.

\vspace{1mm}
\noindent \textbf{Feature Caching and Reuse Mechanisms.} Feature caching, as a relatively popular research direction in recent years, has attracted attention due to its characteristics of training-free and effectiveness. Feature caching~\cite{ma2024deepcache} was initially introduced in the diffusion model based on U-Net, and then applied to solve the timestep redundancy of DiTs~\cite{chen2024delta-dit, selvaraju2024fora}. Early static cache methods focused on the implementation based on scheduling~\cite{ma2024l2c} for feature caching with timestep and model layer customization, exploring redundancy calculations in spatial~\cite{zou2024accelerating, Liu2025SpeCa, Zheng2025Compute}, temporal~\cite{zou2024DuCa}, and conditional dimensions~\cite{yuan2024ditfastattn}. With the introduction of dynamic feature~\cite{Liu2025SpeCa, liu2024timestep, kahatapitiyaAdaptiveCachingFaster2024} update methods, this sample-adaptive feature caching further expanded its effectiveness. Furthermore, with the introduction of predictive caching by TaylorSeer~\cite{liuTaylorSeer2025}, recent work~\cite{fengHiCache2025, zhengFoCa2025,liu2025freqcaacceleratingdiffusionmodels} has explored caching predictions based on different polynomial methods to achieve approximate estimation of features, thereby realizing higher acceleration in diffusion inference.

While predictive methods achieve \textit{impressive acceleration}, they often suffer from \textit{poor consistency} with the original image or video. Large caching intervals without correction easily distort spatial structures, as \textit{simple polynomial methods} fail to capture the \textit{complex, coupled dynamics} of diffusion. These limitations motivate a \textit{learnable} predictor that models temporal evolution more accurately through a \textit{stage-wise} design tailored to different diffusion phases.

\subsection{Kolmogorov-Arnold Network} 
Kolmogorov–Arnold Network (KAN)\cite{Liu2024KANKN} is a neural architecture grounded in the Kolmogorov–Arnold representation theorem, which states that \textit{any multivariate continuous function can be expressed through compositions and summations of univariate functions}. Unlike MLP with fixed activations, KAN uses learnable spline-based univariate functions at each neuron, providing flexible nonlinear modeling. Subsequent studies~\cite{vaca2024kolmogorov, genet2024temporal, livieris2024c} have demonstrated their effectiveness in diverse time-series prediction tasks, confirming their ability to capture complex temporal relationships. With strong functional expressiveness and adaptive approximation capability, KAN serves as an ideal learnable predictor for modeling the nuanced dynamics of the diffusion process.

These insights motivate us to adopt KAN as a learnable predictor to correct the drift present in training-free caching. When integrated with our stage-aware modeling of diffusion dynamics, this yields a learnable-cache paradigm for accelerating diffusion inference.

\begin{figure*}[hbtp]
    \centering
    \includegraphics[width=\linewidth]
    {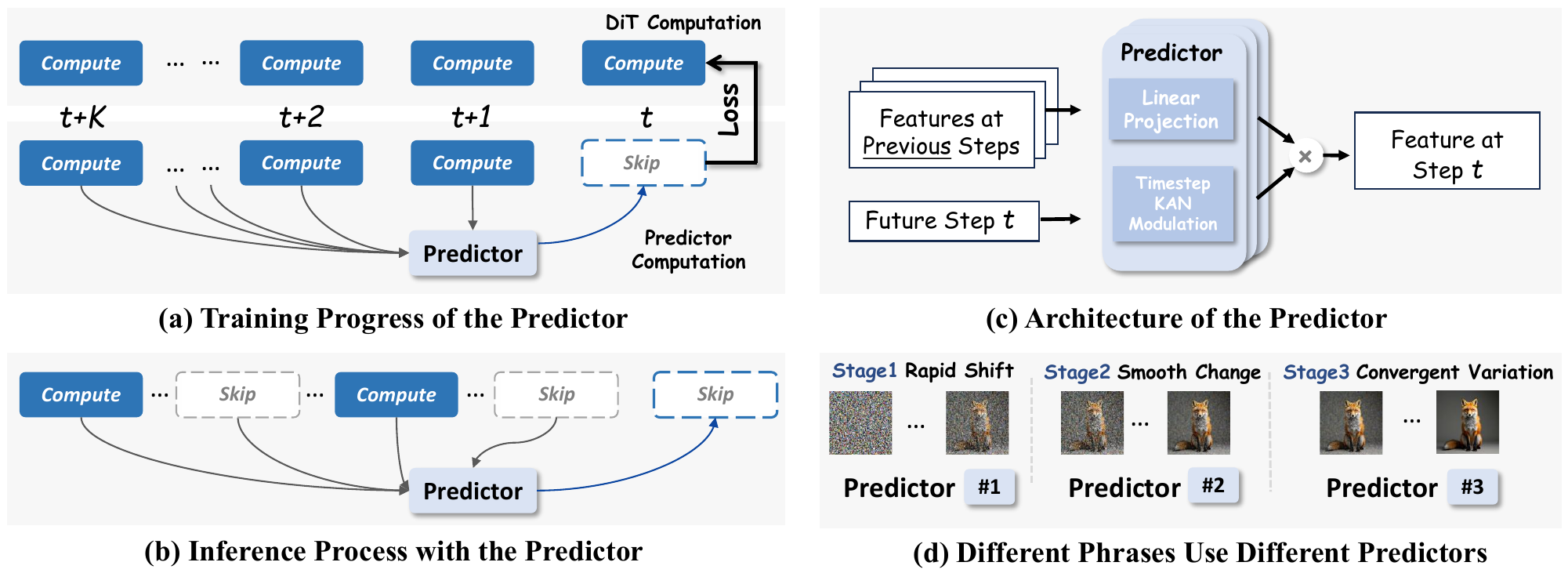}
    \vspace{-7mm}
    \caption{\textbf{Overview of the training-based stage-aware learnable predictor framework.} \textbf{(a) Training progress} of the \textit{LESA} uses outputs from the previous $K$ steps as input and updates the predictor with the DiT activation output. \textbf{(b) Inference process} with the learned predictor skips part of the DiT computations by predicting features. \textbf{(c) Predictor architecture} processes features from previous steps with a linear projection and processes the timestep with a KAN module, whose outputs are multiplied to generate the feature at step $t$. \textbf{(d) Stage-aware denoising} divides the trajectory into three stages and assigns a dedicated predictor to each stage.
}
    \label{fig:method}
    \vspace{-4mm}
\end{figure*}

\section{Method}
\label{sec:method}
\vspace{-1mm}

\subsection{Preliminary}

\noindent \textbf{Feature Caching in Diffusion Models}
Feature caching aims to accelerate diffusion inference by avoiding redundant computation across adjacent timesteps. In the conventional \textit{``cache-then-reuse''} paradigm, features computed at timestep $t$ are directly stored and reused for subsequent steps, i.e.,
\begin{equation}
\mathcal{F}(x_{t-k}^l) \approx \mathcal{F}(x_t^l), \quad k \in {1, \dots, N-1}
\end{equation}
This naive reuse offers an ($\mathcal{N}{-}1$)-fold theoretical speedup but neglects temporal feature evolution, leading to rapid error accumulation as $\mathcal{N}$ grows. To mitigate error accumulation, the recent \textit{``cache-then-forecast''} paradigm reformulates feature caching as a forecasting problem, exploiting the smooth and continuous nature of feature trajectories in diffusion models. Instead of copying features, it models temporal feature evolution using finite differences or polynomial extrapolation. The predicted feature at timestep $t-k$ can be expressed as a truncated Taylor expansion:
\begin{equation}
\mathcal{F}_{\text{pred},m}(x_{t-k}^l) = \mathcal{F}(x_t^l) + \sum_{i=1}^{m}\frac{\Delta^i \mathcal{F}(x_t^l)}{i! \cdot N^i}(-k)^i,
\end{equation}
where $\Delta^i \mathcal{F}(x_t^l)$ denotes the $i$-th finite difference. This formulation transforms caching from simple reuse to temporal prediction, enabling accurate feature estimation.


\noindent \textbf{Kolmogorov-Arnold Network}
KAN is built upon the Kolmogorov–Arnold representation theorem, which states that
any multivariate continuous function can be expressed using only univariate
functions composed with linear projections.
Following this principle, a generic KAN layer realizes the mapping
$\mathbf{x} \mapsto \mathbf{y}$ in the form
\begin{equation}
\mathbf{y} 
\;=\; 
\sum_{m=1}^{M} \mathbf{w}_m\, \phi_m\!\big(\mathbf{a}_m^{\top}\mathbf{x}\big),
\label{eq:kan}
\end{equation}
where $\mathbf{a}_m$ and $\mathbf{w}_m$ parameterize the inner and outer linear
mappings, and each $\phi_m(\cdot)$ is a learnable spline-based univariate
function.
Compared with networks using fixed nonlinearities, KAN explicitly represents
functions as combinations of such univariate components, which improves its
ability to approximate complex smooth dynamics.

\subsection{LESA Framework}

\noindent \textbf{Stage-Aware Temporal Segmentation}
The dynamics of the diffusion denoising trajectory are inherently non-uniform: As shown in Fig.~\ref{fig:method} (d), the high-noise regime is dominated by coarse feature formation, whereas the low-noise regime emphasizes fine-grained detail refinement. To accurately capture these stage-dependent dynamics, we divide the denoising process into discrete stages, each handled by a specialized sub-model tailored to its distinctive behavior.
Furthermore, the temporal window $K$ is likewise adapted to each stage. A shorter window ($K{=}4$) is employed in the high-noise phase to accommodate its rapid and weakly correlated transitions, whereas a longer window ($K{=}8$) is utilized in the subsequent, more stable phases to better capture the intricate, history-dependent evolution of features.

This stage-aware design enables the \textit{LESA} to leverage specialized temporal modeling strategies that align with the statistical and structural properties of each generation stage.

\noindent \textbf{KAN-based Temporal Modeling}
To model the nonlinear temporal dependency from cached features 
$\{\mathbf{h}_{t+K-1}, \mathbf{h}_{t+K-2}, \ldots, \mathbf{h}_{t}\}$ 
to the target feature $\mathbf{h}_{t-1}$,
we design a \textit{compact residual LESA predictor} 
that combines a linear feature projection with a 
Kolmogorov–Arnold Network (KAN) for timestep modulation as illustrated in Fig.~\ref{fig:method} (c).
Specifically, the cached feature sequence is linearly projected to obtain a feature representation:
\begin{equation}
\mathbf{z} 
= \mathbf{W}
[\mathbf{h}_{t+K-1}, \mathbf{h}_{t+K-2}, \ldots, \mathbf{h}_{t}]
+ b,
\label{eq:feat_linear}
\end{equation}
where $\mathbf{W}$ and $b$ are learnable parameters, and $[\cdot]$ denotes concatenation. Meanwhile, we compute the relative temporal offsets between the prediction step 
and all timesteps within the full sequence of $L$ timesteps as 
$\Delta t_i = t_{\mathrm{pred}} - t_i$, for $i = L - 1, \ldots, 0$, 
and feed them into the KAN to produce a scalar temporal modulation factor:
\begin{equation}
\alpha = f_{\mathrm{KAN}}(\Delta t_{L-1}, \ldots, \Delta t_0).
\label{eq:kan_scalar}
\end{equation}

Let $\Delta \mathbf{t} = [\Delta t_{L-1}, \ldots, \Delta t_0]^\top$ be the relative timestep vector.
Following the general KAN form in Eq.~\ref{eq:kan}, we write
\begin{equation}
\alpha 
= f_{\mathrm{KAN}}(\Delta \mathbf{t})
= \sum_{m=1}^{M} w_m \, \phi_m\big(a_m^\top \Delta \mathbf{t}\big),
\label{eq:kan_alpha_decomp}
\end{equation}

where $a_m$ and $w_m$ are learnable coefficients, and each $\phi_m(\cdot)$ is a smooth spline-based function. This yields a flexible scalar function $\alpha = g(t_{\mathrm{pred}})$ that modulates the residual update strength: the predictor learns larger $\alpha$ in early high-noise steps and smaller values during late refinement.

The scalar factor is then broadcast-multiplied with the feature to form 
the residual update:
\begin{equation}
\hat{\mathbf{h}}_{t-1}
\;=\;
\mathbf{h}_{t}
\;+\;
\alpha \, \mathbf{z},
\label{eq:backward_res}
\end{equation}
where $\alpha$ modulates all feature dimensions uniformly across channels.
This formulation follows a separation-of-variables principle:
The linear projection transforms the feature sequence in space,
while the KAN models a continuous scalar temporal modulation over the entire $L$-step sequence.


\noindent \textbf{Training Procedure.}
Our training procedure consists of three stages: \textit{Data Preparation}, \textit{Ground-Truth Guided Training}, \textit{Closed-Loop Autoregressive Training}.

We first run the base diffusion model without acceleration and cache its trajectories. For each prompt, we save the initial noise and the feature at all denoising timesteps.

In the first stage, \textit{LESA} is trained under a \textbf{\textit{ground-truth guided regime}}. At each timestep $t$, as illustrated in Fig.~\ref{fig:method}(a), it receives the past $K$ GT features along with the timestep embedding, and is optimized by minimizing the $L_1$ loss between its prediction and the GT feature at timestep $t$.

In the second stage, we adopt a \textbf{\textit{closed-loop autoregressive training scheme}} that mirrors real inference. The predictor now conditions on the past $K$ features composed of both its own historical predictions and the model’s actual outputs. This exposure to accumulated drift enables the predictor to learn how features evolve under imperfect inputs, resulting in more robust behavior during accelerated inference.

\textit{For a comprehensive description of the training details, please refer to the supplementary material.}

\section{Experiments}
\label{sec:experiments}

\noindent \textbf{Evaluation and Metrics}
For the text-to-image task, we evaluate on DrawBench\citep{saharia2022drawbench} using ImageReward\citep{xu2023imagereward} and CLIP Score\citep{hessel2021clipscore} as quality metrics, and PSNR, SSIM\citep{wang2004imagequality}, and LPIPS\citep{zhangUnreasonableEffectivenessDeep2018} as perceptual metrics.
For the text-to-video task, we use the same perceptual metrics to assess frame-wise fidelity. We highlight the \textbf{best results in bold} and \underline{underline our method when it ranks second}.
\textit{See the supplementary material for configuration and evaluation details, along with additional experiments on distilled models.}

\begin{table*}[hbt]
\centering
\caption{\centering
\textbf{Quantitative comparison in text-to-image generation} for FLUX.1-schnell.
}
\vspace{-3mm}
  \resizebox{\textwidth}{!}{ 
    \begin{tabular}{l | c  c | c  c | c c  | c c c}
        \toprule
        {\bf Method}
        & \multicolumn{4}{c|}{\textbf{Acceleration}} 
        & \multicolumn{2}{c|}{\textbf{Quality Metrics}} 
        & \multicolumn{3}{c}{\textbf{Perceptual Metrics}}\rule{0pt}{2ex}\\
        \cline{2-10}
        {\bf FLUX}
        & \textbf{Latency(s) \(\downarrow\)} 
        & \textbf{Speed \(\uparrow\)} 
        & \textbf{FLOPs(T) \(\downarrow\)}  
        & \textbf{Speed \(\uparrow\)} 
        & \textbf{\makecell{Image\\Reward\(\uparrow\)}}
        & \textbf{\makecell{CLIP\\Score\(\uparrow\)}}
        & \textbf{PSNR\(\uparrow\)} 
        & \textbf{SSIM\(\uparrow\)}
        & \textbf{LPIPS\(\downarrow\)}\rule{0pt}{2ex}\\
        \midrule

\textbf{[dev]: 50 steps}
& 23.24 & 1.00$\times$ & 3726.87 & 1.00$\times$ & 0.99 & 32.64 & $\infty$ & 1.00 & 0.00  \\


\textbf{$50\%$ steps}
& 11.82 & 1.97$\times$ & 1863.44 & 2.00$\times$ & 0.97 & 32.57 & 29.56 & 0.73 & 0.31  \\

\textbf{PAB}
& 17.84 & 1.30$\times$ & 3013.13 & 1.24$\times$ &0.95 & 32.55 & 28.84 & 0.67 & 0.40 \\

\textbf{DBCache}
& 16.88 & 1.38$\times$ & 2384.29 & 1.56$\times$ & 1.01 & 32.53 & 33.86 & 0.87 & 0.12 \\

\midrule

\textbf{FORA} ($\mathcal{N}$=3) 
& 9.06 & 2.57$\times$ & 1267.89 & 2.94$\times$ & 0.99 & 32.27 & 30.65 & 0.77 & 0.25 \\

\textbf{TeaCache} ($l$=0.6) 
& 9.13 & 2.55$\times$ & 1342.20 & 2.78$\times$ & 0.91 & 32.11 & 29.03 & 0.68 & 0.40  \\

\textbf{TaylorSeer} ($\mathcal{N}$=4, $O$=2)
& 8.78 & 2.64$\times$ & 1118.86 & 3.57$\times$ & \textbf{1.02} & 32.41 & 29.87 & 0.73 & 0.30  \\

\rowcolor{gray!20}
\textbf{LESA} ($\mathcal{N}$=5) 
& \textbf{6.57} & \textbf{3.54}$\times$ & \textbf{969.11} & \textbf{3.85}$\times$ & \underline{1.00} & \textbf{32.70} & \textbf{30.96} & \textbf{0.77} & \textbf{0.24}\\

\midrule

\textbf{FORA}($\mathcal{N}$=5) {\textcolor{red}{$^{\dagger}$}}
& 5.97 & 3.89$\times$ & 820.80 & 4.54$\times$ & 0.82 & 32.48 & 28.44 & 0.60 & 0.50  \\ 

\textbf{\texttt{ToCa}}($\mathcal{N}$=8, $\mathcal{R}$=75\%) 
& 12.39 & 1.88$\times$ & 829.86 & 4.49$\times$ & 0.95 & 32.60 & 29.07 & 0.64 & 0.43   \\ 

\textbf{\texttt{DuCa}}($\mathcal{N}$=8, $\mathcal{R}$=70\%) 
& 9.40 & 2.47$\times$ & 858.27 & 4.34$\times$ & 0.94 & 32.58 & 29.06 & 0.64 & 0.43   \\ 

\textbf{TeaCache}($l$=1.0) {\textcolor{red}{$^{\dagger}$}}
& 7.07 & 3.29$\times$ & 820.55 & 4.54$\times$ & 0.84 & 31.88 & 28.61 & 0.64 & 0.48  \\

\textbf{TaylorSeer}($\mathcal{N}$=6, $O$=2)
& 6.73 & 3.45$\times$ & 746.28 & 4.99$\times$ & \textbf{1.02} & 32.53 & 28.94 & 0.66 & 0.40  \\

\rowcolor{gray!20}
\textbf{LESA} ($\mathcal{N}$=7) 
& \textbf{5.19} & \textbf{4.48}$\times$ & \textbf{745.51} & \textbf{5.00}$\times$ & \underline{0.98} & \textbf{32.88} & \textbf{30.17} & \textbf{0.72} & \textbf{0.32}\\

\midrule

\textbf{FORA}($\mathcal{N}$=7) {\textcolor{red}{$^{\dagger}$}}
& 5.09 & 4.57$\times$ & 597.25 & 6.24$\times$ & 0.68 & 31.90 & 28.32 & 0.59 & 0.54  \\ 

\textbf{\texttt{ToCa}}($\mathcal{N}$=12, $\mathcal{R}$=85\%) {\textcolor{red}{$^{\dagger}$}}
& 9.82 & 2.37$\times$ & 618.57 & 6.02$\times$ & 0.80 & 32.32 & 28.70 & 0.60 & 0.52   \\ 

\textbf{\texttt{DuCa}}($\mathcal{N}$=12, $\mathcal{R}$=80\%) {\textcolor{red}{$^{\dagger}$}}
& 7.74 & 3.00$\times$ & 646.97 & 5.76$\times$ & 0.77 & 32.20  & 28.71 & 0.60 & 0.53   \\ 

\textbf{TeaCache}($l$=1.4) {\textcolor{red}{$^{\dagger}$}}
& 6.14 & 3.79$\times$ & 671.51 & 5.55$\times$ & 0.74 & 31.78 & 28.12 & 0.48 & 0.68  \\

\textbf{TaylorSeer}($\mathcal{N}$=9, $O$=2) {\textcolor{red}{$^{\dagger}$}}
& 5.85 & 3.97$\times$ & 597.25 & 6.24$\times$ & 0.86 & 32.04  & 28.38 & 0.59 & 0.51  \\

\rowcolor{gray!20}
\textbf{LESA} ($\mathcal{N}$=10) 
& \textbf{4.28} & \textbf{5.43}$\times$ & \textbf{596.44} & \textbf{6.25}$\times$ & \textbf{0.91} & \textbf{32.65} & \textbf{29.65} & \textbf{0.67} & \textbf{0.40}\\

\midrule
\textbf{[schnell]: 4 steps} 
& 2.13 & 1.00$\times$ & 278.41 & 1.00$\times$ & 0.93 & 34.09 & $\infty$ & 1.00 & 0.00 \\
\rowcolor{gray!20}
\textbf{LESA} ($\mathcal{N}$=3) 
& 1.31 & 1.63$\times$ & 139.21& 2.00$\times$& 0.95& 34.48& 30.31& 0.80& 0.16\\
\rowcolor{gray!20}
\textbf{LESA} ($\mathcal{N}$=4) 
& 1.15 & 1.85$\times$ & 69.61& 4.00$\times$& 0.91& 34.67& 29.21& 0.69& 0.30\\

\bottomrule
\end{tabular}
}

\label{table:FLUX-Metrics}
\raggedright
{
\begin{itemize}
\item\textcolor{red}{$\dagger$} Methods exhibit significant degradation in image quality. 
\end{itemize}
}
\vspace{-6mm}
\end{table*}


\subsection{Text-to-Image Generation}

\subsubsection{FLUX.1-dev and FLUX.1-schnell}
\begin{figure}
    \centering
    \includegraphics[width=\linewidth]
    {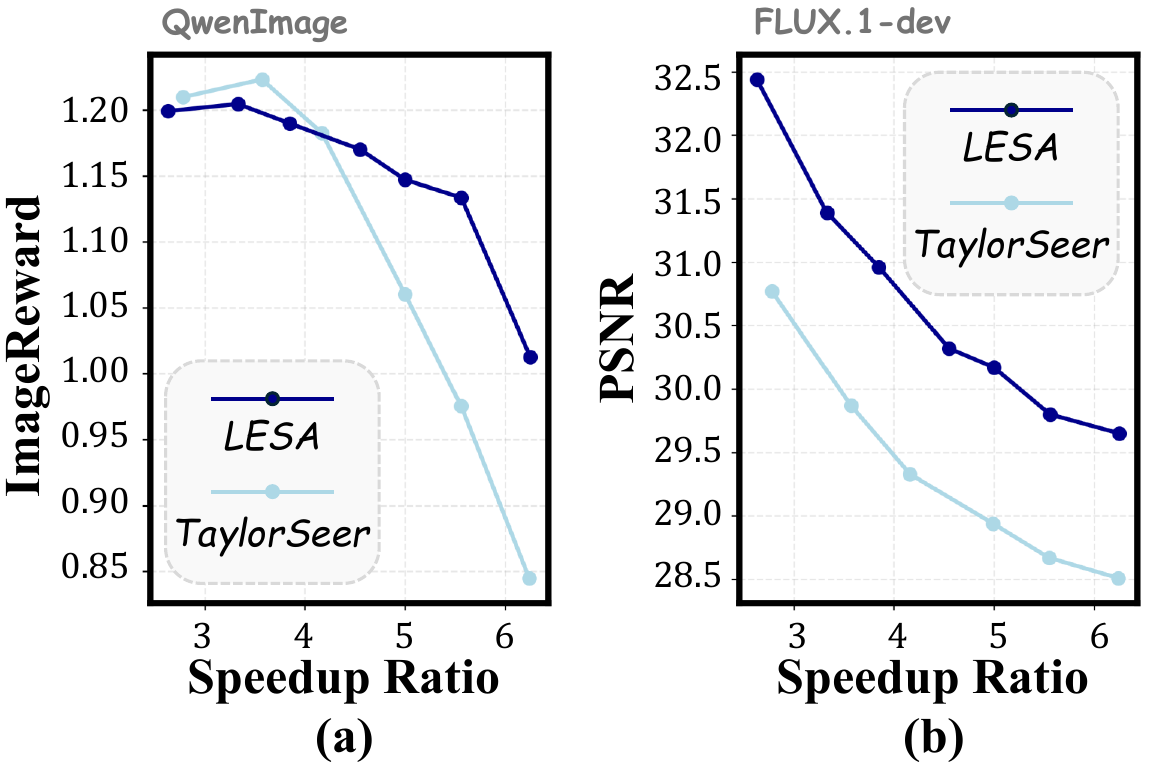}
    \vspace{-8mm}
    \caption{\textbf{Comparison of \textit{LESA} and TaylorSeer under different speedup ratios.} \textit{LESA} preserves better quality metrics and perceptual metrics On QwenImage and FLUX.}
    \label{fig:flux-dev and qwenimage}
    \vspace{-6mm}
\end{figure}

\noindent \textbf{Quantitative Study}
\begin{figure}[hbtp]
    \centering
    \includegraphics[width=\linewidth]
    {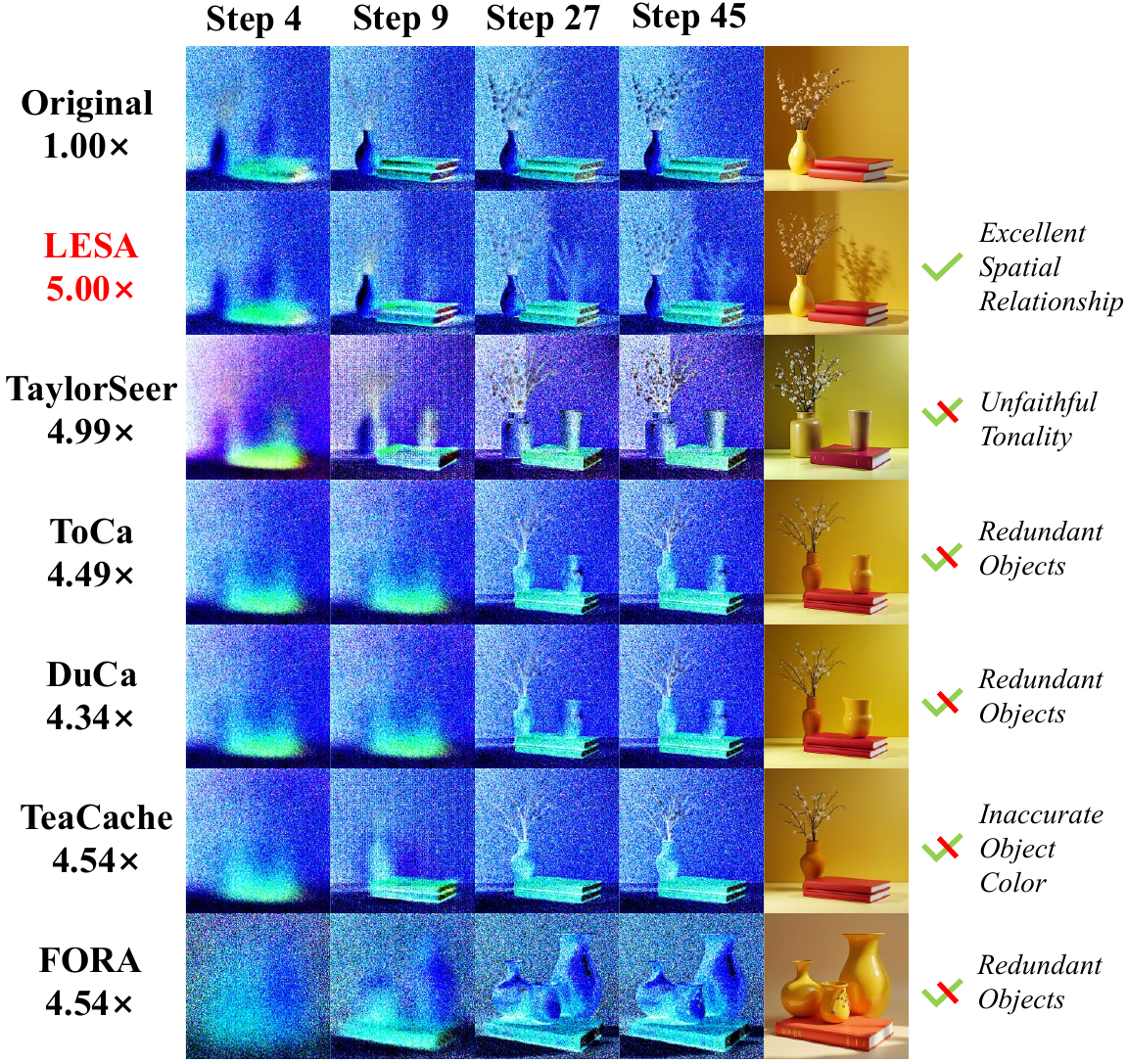}
    \vspace{-6mm}
    \caption{\textbf{Comparison of Sampling Process Image and Final Image.} On FLUX, \textit{LESA} demonstrates significant speedup ratio with excellent spatial organization and accurate color representation.}
    \label{fig:flux-dev}
    \vspace{-4mm}
\end{figure}
We evaluate \textit{LESA} against existing acceleration methods under comparable settings (shown in Tab.~\ref{table:FLUX-Metrics}). In the moderate-acceleration regime, \textit{LESA} ($\mathcal{N}{=}5$) achieves a 3.85$\times$ speedup with a PSNR of 30.96, outperforming TaylorSeer ($\mathcal{N}{=}4$, $O{=}2$) and TeaCache ($l{=}0.6$) in both efficiency and perceptual quality. As acceleration increases, \textit{LESA} maintains a 5.00$\times$ speedup with a CLIP Score of 32.88 and PSNR of 30.17. Even at the highest acceleration, \textit{LESA} sustains a 6.25$\times$ speedup with an ImageReward of 0.91, PSNR of 29.65, and LPIPS of 0.40, demonstrating strong robustness against quality collapse.

Furthermore, on FLUX.1-schnell, \textit{LESA} maintains strong performance even under extreme step reduction like $\mathcal{N}{=}4$, achieving substantial acceleration with only a modest quality drop. Notably, when distilled models behave suboptimally, the cache-based compression of intermediate representations can even improve stability and visual fidelity, further highlighting the practicality of our approach.

\noindent \textbf{Qualitative Study}
As shown in Fig.~\ref{fig:flux-dev}, \textit{LESA} achieves a 5.00$\times$ speedup while preserving spatial structure, color consistency, and alignment with the target image throughout sampling. In contrast, TaylorSeer shows drift and structural distortion despite similar acceleration, and other methods generate inconsistent objects that break semantic coherence.

\subsubsection{Qwen-Image and Qwen-Image-Lightning}

\begin{table*}[htb]
\centering
\caption{\centering
\textbf{Quantitative comparison in text-to-image generation} for Qwen-Image and Qwen-Image-Lightning.}
\vspace{-3mm}
  \resizebox{\textwidth}{!}{ 
    \begin{tabular}{l | c  c | c  c | c  c | c c c}
        \toprule
        {\bf Method}
        & \multicolumn{4}{c|}{\textbf{Acceleration}} 
        & \multicolumn{2}{c|}{\textbf{Quality Metrics}} 
        & \multicolumn{3}{c}{\textbf{Perceptual Metrics}}\rule{0pt}{2ex}\\
        \cline{2-10}
        {\bf Qwen}
        & \textbf{Latency(s) \(\downarrow\)} 
        & \textbf{Speed \(\uparrow\)} 
        & \textbf{FLOPs(T) \(\downarrow\)}  
        & \textbf{Speed \(\uparrow\)} 
        & \textbf{\makecell{Image\\Reward\(\uparrow\)}}
        & \textbf{\makecell{CLIP\\Score\(\uparrow\)}}
        & \textbf{PSNR\(\uparrow\)} 
        & \textbf{SSIM\(\uparrow\)}
        & \textbf{LPIPS\(\downarrow\)}\rule{0pt}{2ex}\\
        \midrule

\textbf{$50$ steps}
& 127.40 & 1.00$\times$ & 12917.56 & 1.00$\times$ & 1.25 & 35.59 & $\infty$ & 1.00 & 0.00 \\


\textbf{$20\%$ steps} {\textcolor{red}{$^{\dagger}$}}
& 25.92 & 4.89$\times$ & 2583.51 & 5.00$\times$ & 0.94 & 34.95 & 28.59 & 0.61 & 0.52 \\

\midrule

\textbf{FORA}($\mathcal{N}$=4){\textcolor{red}{$^{\dagger}$}}
& 38.43 & 3.32$\times$ & 3359.99 & 3.84$\times$ & 0.93 & 34.40 & 28.66 & 0.59 & 0.51 \\

\textbf{\texttt{ToCa}}($\mathcal{N}$=8, $\mathcal{R}$=75\%)
& 61.37 & 2.08$\times$ & 2991.34 & 4.32$\times$ & 1.02 & 34.96 & 28.93 & 0.63 & 0.44 \\

\textbf{\texttt{DuCa}}($\mathcal{N}$=9, $\mathcal{R}$=80\%){\textcolor{red}{$^{\dagger}$}}
& 34.73 & 3.67$\times$ & 2958.13 & 4.37$\times$ & 0.77 & 34.62 & 28.45 & 0.58 & 0.55 \\

\textbf{TaylorSeer}($\mathcal{N}$=6, $O$=2)
& 30.29 & 4.21$\times$ & 2585.46 & 5.00$\times$ & 1.01 & 34.71 & 28.58 & 0.62 & 0.46 \\

\rowcolor{gray!20}
\textbf{LESA} ($\mathcal{N}$=7) 
& \textbf{27.34} & \textbf{4.66}$\times$ & \textbf{2584.90} & \textbf{5.00}$\times$ & \textbf{1.15} & \textbf{35.36} & \textbf{30.18} & \textbf{0.78} & \textbf{0.25}\\

\midrule

\textbf{FORA}($\mathcal{N}$=6){\textcolor{red}{$^{\dagger}$}}
& 28.69 & 4.44$\times$ & 2326.74 & 5.55$\times$ & 0.48 & 33.34 & 28.48 & 0.55 & 0.59 \\

\textbf{\texttt{ToCa}}($\mathcal{N}$=12, $\mathcal{R}$=85\%){\textcolor{red}{$^{\dagger}$}}
& 50.95 & 2.50$\times$ & 2406.20 & 5.37$\times$ & 0.55 & 34.08 & 28.69 & 0.57 & 0.53 \\

\textbf{\texttt{DuCa}}($\mathcal{N}$=12, $\mathcal{R}$=90\%){\textcolor{red}{$^{\dagger}$}}
& 28.57 & 4.46$\times$ & 2171.56 & 5.95$\times$ & 0.41 & 33.38 & 28.38 & 0.57 & 0.60 \\

\textbf{TaylorSeer}($\mathcal{N}$=8, $O$=2){\textcolor{red}{$^{\dagger}$}}
& 25.24 & 5.05$\times$ & 2068.86 & 6.24$\times$ & 0.84 & 33.77 & 28.14 & 0.47 & 0.68 \\

\rowcolor{gray!20}
\textbf{LESA} ($\mathcal{N}$=10) 
& \textbf{22.40} & \textbf{5.69}$\times$ & \textbf{2068.03} & \textbf{6.25}$\times$ & \textbf{1.01} & \textbf{34.93} & \textbf{29.23} & \textbf{0.73} & \textbf{0.34}\\

\midrule
\textbf{[Lightning]: 8 steps} 
& 4.37 & 1.00$\times$ & 560.96 & 1.00$\times$ & 1.29 & 35.32 & $\infty$ & 1.00 & 0.00 \\

\rowcolor{gray!20}
\textbf{LESA} ($\mathcal{N}$=2) 
& 2.95 & 1.48$\times$ & 350.61& 1.60$\times$ & 1.28& 35.20& 33.42& 0.85& 0.10\\

\rowcolor{gray!20}
\textbf{LESA} ($\mathcal{N}$=3) 
& 2.48 & 1.76$\times$ & 280.49& 2.00$\times$ & 1.27& 35.24& 31.38& 0.80& 0.16\\


\rowcolor{gray!20}
\textbf{LESA} ($\mathcal{N}$=4)
& 1.52 & 2.88$\times$ & 140.25 & 4.00$\times$ & 1.23 & 35.66 & 28.85 & 0.63 & 0.37\\

\bottomrule
\end{tabular}
}

\label{table:qwen-image-Metrics}
\raggedright
{
\begin{itemize}
\item\textcolor{red}{$\dagger$} Methods exhibit significant degradation in image quality. 
\end{itemize}
}
\vspace{-2mm}
\end{table*}

\begin{figure*}[htbp]
    \centering
    \includegraphics[width=\linewidth]
    {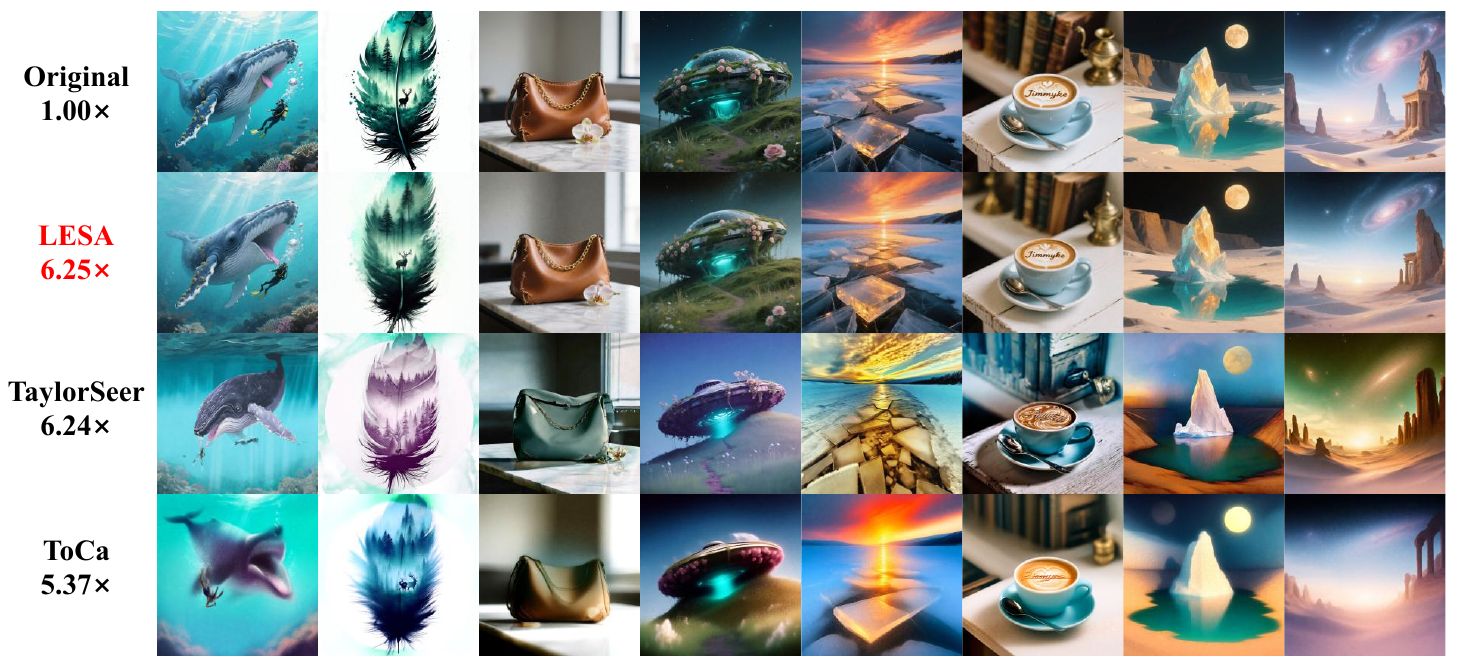}
    \vspace{-8mm}
    \caption{\textbf{Qualitative Comparison of Generated Images.} On Qwen-Image, \textit{\textit{LESA}} delivers higher speedup with high image quality.}
    \label{fig:qwen-image}
    \vspace{-4mm}
\end{figure*}
\noindent \textbf{Quantitative Study}
We evaluate our approach on Qwen-Image and its distilled variant, Qwen-Image-Lightning. As shown in Tab.~\ref{table:qwen-image-Metrics}, \textit{LESA} ($\mathcal{N}{=}7$) achieves a 5.00$\times$ speedup with\textbf{ high image consistency}, clearly outperforming other methods that typically fall below 30 PSNR and above 0.40 LPIPS. With a stronger acceleration setting, \textit{LESA} ($\mathcal{N}{=}10$) reaches 6.25$\times$ speedup while still maintaining competitive fidelity, whereas baseline methods experience significant quality degradation. On Qwen-Image-Lightning, \textit{LESA}  achieves up to 4$\times$ speedup while consistently preserving high image quality. Even at extreme step setting, it remains close to the original baseline in perceptual fidelity. 

\vspace{1mm}
\noindent \textbf{Qualitative Study}
Across the visual comparison in Fig.~\ref{fig:qwen-image}, \textit{LESA} at 6.25$\times$ delivers results closest to the original image. It preserves whale anatomy and lighting, and clearly retains the latte foam lettering, while other methods exhibit global appearance shifts and noticeable loss of details.

\subsection{Text-to-Video Generation}
\begin{table*}[htb]
\centering
\vspace{-4mm}
\caption{\centering
\textbf{Quantitative comparison of text-to-video generation} on HunyuanVideo.}
\vspace{-2mm}
\setlength\tabcolsep{5.0pt} 
\small
\resizebox{\textwidth}{!}{
\begin{tabular}{l | c c c c | c c c}
    \toprule
    {\bf Method} 
    &\multicolumn{4}{c|}{\bf Acceleration} 
    &\multicolumn{3}{c}{\textbf{Perceptual Metrics}}\rule{0pt}{2ex}\\
    \cline{2-8}
    {\bf HunyuanVideo}
    & {\bf Latency(s) $\downarrow$}
    & {\bf Speed $\uparrow$}
    & {\bf FLOPs(T) $\downarrow$}
    & {\bf Speed $\uparrow$} 
    & \textbf{PSNR\(\uparrow\)} 
    & \textbf{SSIM\(\uparrow\)} 
    & \textbf{LPIPS\(\downarrow\)}\rule{0pt}{2ex}\\ 
    \midrule

\textbf{Original: 50 steps} 
& 98.91 & {1.00}$\times$ & 29773.0 & {1.00}$\times$ & $\infty$ & 1.00 & 0.00 \\

\midrule
  
$22\%$ \textbf{steps}  
& 22.98 & {4.30}$\times$ & 6550.1 & 4.55$\times$ & 17.65 & 0.59 & 0.42 \\

\textbf{\texttt{ToCa}}($\mathcal{N}=5$, $\mathcal{R}=90\%$) 
& 26.12 & 3.79$\times$ & 7006.2 & 4.25$\times$ & 17.04 & 0.54 & 0.44 \\

\textbf{\texttt{DuCa}}($\mathcal{N}=5$, $\mathcal{R}=90\%$) 
& 23.08 & 4.29$\times$ & 6483.2 & 4.48$\times$ & 17.08 & 0.54 & 0.43\\

\textbf{TeaCache}($l=0.4$) 
& 21.83 & 4.53$\times$ & 6550.1 & 4.55$\times$  & 18.25 & 0.61 & 0.38\\

\textbf{FORA}($N=5$) 
& 22.61 & 4.37$\times$ & 5960.4 & 5.00$\times$ & 17.00 & 0.53 & 0.44 \\

\textbf{TaylorSeer}($\mathcal{N}=5$, $O=1$) 
& 23.66 & 4.18$\times$ & 5960.4 & 5.00$\times$ & 17.29 & 0.55 & 0.42\\

\textbf{SpeCa}($\mathcal{N}_{\text{max}}=8$, $\mathcal{N}_{\text{min}}=2$) 
& 23.22 & 4.26$\times$ & 5692.7 & 5.23$\times$ & 17.73 & 0.59 & 0.39\\

\textbf{ClusCa}($\mathcal{N}=5$, $O=1$, $K=16$) 
& 24.35 & 4.06$\times$ & 5373.0 & 5.54$\times$ & 17.28 & 0.55 & 0.42\\

\rowcolor{gray!20}
\textbf{LESA($\mathcal{N}=7$)} 
& 19.37 & {5.11}$\times$ & 5960.4 & {5.00}$\times$ & \textbf{21.43} & \textbf{0.72} & \textbf{0.29} \\

\rowcolor{gray!20}
\textbf{LESA($\mathcal{N}=8$)} 
& \textbf{17.47} & \textbf{5.66}$\times$ & \textbf{5354.9} & \textbf{5.56}$\times$ & {21.05} & {0.70} & {0.32} \\

\bottomrule
\end{tabular}}
\label{table:HunyuanVideo-Metrics}
\vspace{-2mm}
\end{table*}

\noindent \textbf{Quantitative Study}
As shown in Tab.~\ref{table:HunyuanVideo-Metrics}, directly reducing the number of steps brings notable speedup but results in a substantial degradation of perceptual quality, and existing baselines exhibit a similar trade-off. In contrast, \textit{LESA} simultaneously achieves faster inference and markedly better visual fidelity. With $\mathcal{N}{=}7$, it attains the best overall performance, improving PSNR by \textbf{23.9\%} over TaylorSeer; even under the more aggressive $\mathcal{N}{=}8$ setting, it still provides the highest acceleration while retaining strong perceptual quality, with a \textbf{21.7\%} PSNR gain.

\vspace{1mm}
\noindent \textbf{Qualitative Study}
\begin{figure*}[htbp]
    \centering
    \includegraphics[width=\linewidth]
    {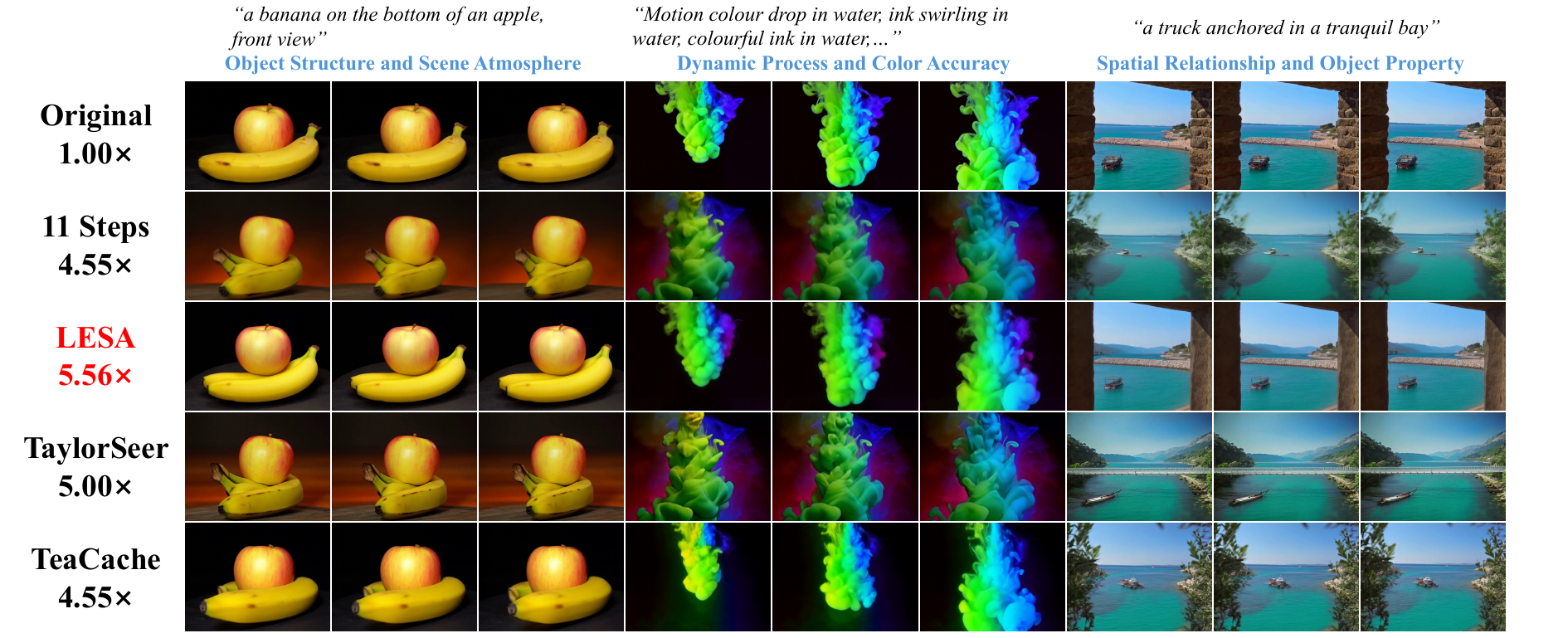}
    \vspace{-6mm}
    \caption{\textbf{Qualitative Comparison of Early, Middle, and Late Frames From Generated Videos.} On HunyuanVideo, \textit{LESA} constructs continuous dynamic process with excellent subject accuracy under high speedup ratio.}
    \label{fig:hunyuanvideo}
    \vspace{-4mm}
\end{figure*}
As shown in Fig.~\ref{fig:hunyuanvideo}, \textit{LESA} at 5.56$\times$ preserves scene semantics and temporal consistency more reliably than all baselines. For \textit{banana on the bottom of an apple},'' it keeps the object placement stable, while others distort shapes or shift positions. In ``\textit{ink swirling in water},'' it maintains multicolor flow and smooth dynamics, whereas competing methods lose color fidelity or introduce artifacts.

\subsection{Ablation Study}
\begin{table}[t]
\centering
\caption{\textbf{Ablation study for Interval, Stage Segmentation and Timestep Modulation Model} for \textit{LESA} on FLUX}
\vspace{-3mm}
\resizebox{\columnwidth}{!}{%
\begin{tabular}{c c c | c c | c c c}
    \toprule
    \multicolumn{3}{c|}{\bf Experiment Settings} 
    & \multicolumn{2}{c|}{\textbf{Quality Metrics}} 
    & \multicolumn{3}{c}{\textbf{Perceptual Metrics}} \\
    \cline{1-8}
    {\bf Interval} & {\bf Segmented} & {\bf Module}
    & \textbf{\makecell{Image\\Reward\(\uparrow\)}}
    & \textbf{\makecell{CLIP\\Score\(\uparrow\)}}
    & \textbf{PSNR\(\uparrow\)} 
    & \textbf{SSIM\(\uparrow\)}
    & \textbf{LPIPS\(\downarrow\)}\\
    \midrule

$\mathcal{N}$ = 5 & \ding{52} & KAN
& 1.00& 32.70& 30.96& 0.77& 0.24\\

$\mathcal{N}$ = 5 & \ding{56} & KAN
& 1.00& 32.67& 30.77& 0.77& 0.25\\

$\mathcal{N}$ = 5 & \ding{52} & MLP
& 0.99& 32.69& 30.76& 0.77& 0.25\\

$\mathcal{N}$ = 5 & \ding{56} & MLP
& 0.95& 32.50& 30.29& 0.73& 0.31\\

\midrule

$\mathcal{N}$ = 7 & \ding{52} & KAN
& 0.98& 32.88& 30.17& 0.72& 0.32\\

$\mathcal{N}$ = 7 & \ding{56} & KAN
& 0.95& 32.75& 30.10& 0.71& 0.33\\

$\mathcal{N}$ = 7 & \ding{52} & MLP
& 0.96& 32.72& 30.11& 0.71& 0.33\\

$\mathcal{N}$ = 7 & \ding{56} & MLP
& 0.95& 32.65& 30.08& 0.71& 0.34\\

\midrule

$\mathcal{N}$ = 10 & \ding{52} & KAN
& 0.91& 32.65& 29.65& 0.67& 0.40\\

$\mathcal{N}$ = 10 & \ding{56} & KAN
& 0.88& 32.58& 29.60& 0.66& 0.41\\

$\mathcal{N}$ = 10 & \ding{52} & MLP
& 0.89& 32.57& 29.59& 0.66& 0.42\\

$\mathcal{N}$ = 10 & \ding{56} & MLP
& 0.88& 32.51& 29.55& 0.66& 0.43\\

\bottomrule
\end{tabular}
}
\label{table:Ablation-Metrics}
\vspace{-7mm}
\end{table}

As detailed in Tab.~\ref{table:Ablation-Metrics}, our ablation study investigates the impact of the inference interval $\mathcal{N}$, stage segmentation, and the timestep modulation module. The results consistently demonstrate that the combination of stage segmentation with a KAN-based module yields superior performance. This advantage is particularly pronounced at larger intervals, such as $\mathcal{N}{=}10$, which involve more significant temporal dynamics. In contrast, at smaller intervals ($\mathcal{N}{=}5$), the high feature similarity between steps leads to only marginal performance differences among the configurations. Crucially, at $\mathcal{N}{=}10$, our full model not only outperforms the baseline by a significant 3.4\% in ImageReward but also achieves concurrent improvements across all perceptual metrics.

\vspace{-2mm}
\section{Conclusion}
\label{sec:conclusion}
\vspace{-2mm}

To address a key limitation in predictable feature caching,namely that \textit{fixed prediction schemes cannot adapt to the stage-dependent dynamics of diffusion models}, we introduce \textit{LESA}, a \textbf{stage-aware learnable predictor} built on a two-stage training strategy. By partitioning the denoising trajectory into distinct stages and equipping each with a dedicated KAN-based temporal predictor, \textit{LESA} learns stage-specific temporal feature evolution instead of relying on fixed heuristics. Experiments on image and video generation demonstrate that \textit{LESA} maintains higher perceptual quality and competitive performance even at $6.25\times$ acceleration, validating that modeling stage-specific dynamics is an effective approach for efficient diffusion acceleration.

\section*{Acknowledgement}
\mbox{This work was sponsored by WUYING - Alibaba Cloud.}
\vspace{-2mm}
{
    \small
    \bibliographystyle{ieeenat_fullname}
    \bibliography{main}
}

\clearpage
\setcounter{page}{1}
\maketitlesupplementary
\raggedbottom
\section{Method Details}
\subsection{Preliminary}

\noindent \textbf{Diffusion Transformer Architecture.} The Diffusion Transformer (DiT) processes an input $\mathbf{x}_t = \{x_i\}_{i=1}^{H \times W}$ through a hierarchical structure $\mathcal{G} = g_1 \circ g_2 \circ \cdots \circ g_L$. Each block $g_l$, where the superscript $l$ denotes the layer index ranging from $1$ to $L$, is composed as $g_l = \mathcal{F}_{\text{SA}}^l \circ \mathcal{F}_{\text{CA}}^l \circ \mathcal{F}_{\text{MLP}}^l$ and sequentially applies self-attention, cross-attention, and multilayer perceptron components. Critically, these components $\mathcal{F}_{\text{SA}}^l$, $\mathcal{F}_{\text{CA}}^l$, and $\mathcal{F}_{\text{MLP}}^l$ dynamically evolve over time by adapting to the varying noise levels at each denoising timestep. This temporal adaptation allows the model to follow the evolving noise schedule during denoising. Since most adjacent timesteps produce highly similar features, this temporal similarity naturally enables cache-based acceleration.

\subsection{Comparison of KAN and MLP}
\paragraph{From Kolmogorov--Arnold representation to KAN layers.}
Kolmogorov--Arnold Network (KAN) is inspired by the classical
Kolmogorov--Arnold representation theorem, which states that any
continuous multivariate function $f : \mathbb{R}^d \to \mathbb{R}$ can be
expressed using compositions and summations of univariate functions. A
typical abstract form is
\begin{equation}
    f(x_1,\dots,x_d)
    = \sum_{q=1}^{Q}
        \Phi_q\!\left(
            \sum_{p=1}^{d} \psi_{q,p}(x_p)
        \right),
    \label{eq:ka_theorem}
\end{equation}
where each $\Phi_q$ and $\psi_{q,p}$ is a scalar-valued univariate
function.

KAN turns this theoretical representation into a trainable neural
architecture by parameterizing the involved univariate functions with
learnable spline functions. Concretely, a single KAN layer maps
$h^{(\ell-1)}(x)\in\mathbb{R}^{D_{\ell-1}}$ to
$h^{(\ell)}(x)\in\mathbb{R}^{D_{\ell}}$ as
\begin{equation}
    h^{(\ell)}_j(x)
    =
    \sum_{m=1}^{M_\ell}
      w^{(\ell)}_{j,m}\,
      \phi^{(\ell)}_{j,m}\!\big(
        a^{(\ell)\top}_{j,m}\, h^{(\ell-1)}(x)
      \big)
    + b^{(\ell)}_j.
    \label{eq:kan_layer_general}
\end{equation}
Here $j=1,\dots,D_\ell$ denotes the output channel index and $a^{(\ell)}_{j,m} \in \mathbb{R}^{D_{\ell-1}}$ are learnable linear
projections,
$w^{(\ell)}_{j,m}\in\mathbb{R}$ are output mixing weights, and each
$\phi^{(\ell)}_{j,m}:\mathbb{R}\to\mathbb{R}$ is a learnable
spline-based univariate function.
Stacking such layers yields a deep KAN
$f(x)=h^{(L)}(x)$ with $L$ KAN layers.

In many applications we are interested in a scalar modulation function
$\alpha:\mathbb{R}^{d}\to\mathbb{R}$, for instance as a temporal
weighting term depending on a timestep feature vector
$\Delta t\in\mathbb{R}^{d}$.
In this case we can specialize the general layer in
Eq.~\eqref{eq:kan_layer_general} to a \emph{scalar-output}
KAN layer. Let $D_0 = d$, $D_1=1$, and omit the layer index $\ell$ and output
index $j$. Then Eq.~\eqref{eq:kan_layer_general} simplifies to
\begin{equation}
    \alpha_{\mathrm{KAN}}(\Delta t)
    =
    \sum_{m=1}^{M}
      w_{m}\,
      \phi_{m}\!\big(
        a_{m}^\top \Delta t
      \big)
    + b,
    \label{eq:kan_scalar_bias}
\end{equation}
where $M$ is the number of spline components, $a_m\in\mathbb{R}^d$ are
learnable projection vectors, $w_m\in\mathbb{R}$ are learnable outer
weights, and each $\phi_m$ is a learnable univariate spline.

For notational simplicity, the bias $b$ can be absorbed into an
additional constant component.
Define ${\phi}_1(z)\equiv 1$ and
${w}_1 = b$.
Then Eq.~\eqref{eq:kan_scalar_bias} can be rewritten as
\begin{equation}
    \alpha_{\mathrm{KAN}}(\Delta t)
    =
    \sum_{m=1}^{M}
      {w}_{m}\,
      {\phi}_{m}\!\big(
        a_{m}^\top \Delta t
      \big),
\end{equation}
Thus, a scalar KAN layer is a linear combination of learnable
spline-based responses $\phi_m(a_m^\top \Delta t)$, each acting on a
learnable one-dimensional projection of the input.

\vspace{-3mm}
\paragraph{MLP-based scalar mapping.}
For comparison, a standard multilayer perceptron (MLP) with one hidden
layer and a fixed activation function $\sigma$ realizes a scalar mapping
$\alpha_{\mathrm{MLP}}:\mathbb{R}^d\to\mathbb{R}$ in the form
\begin{equation}
    \alpha_{\mathrm{MLP}}(\Delta t)
    = v^\top \sigma(U\Delta t + c),
    \label{eq:mlp_scalar}
\end{equation}
where $U\in\mathbb{R}^{H\times d}$ and $v\in\mathbb{R}^{H}$ are
learnable weights, $c\in\mathbb{R}^{H}$ is a bias, and $H$ is the hidden
width.
Writing out Eq.~\eqref{eq:mlp_scalar} component-wise gives
\begin{equation}
    \alpha_{\mathrm{MLP}}(\Delta t)
    = \sum_{h=1}^{H}
        v_h\,
        \sigma\!\big(
          u_h^\top \Delta t + c_h
        \big),
\end{equation}
where $u_h^\top$ denotes the $h$-th row of $U$.

\vspace{-3mm}
\paragraph{Formula-level comparison and advantages of KAN.}
Equations~\eqref{eq:kan_scalar} and~\eqref{eq:mlp_scalar} reveal a
common structure:
\begin{align}
    \alpha_{\mathrm{MLP}}(\Delta t)
    &= \sum_{h=1}^{H}
        v_h\,
        \underbrace{\sigma\!\big(
          u_h^\top \Delta t + c_h
        \big)}_{\text{fixed-shape nonlinearity}}, \\
    \alpha_{\mathrm{KAN}}(\Delta t)
    &= \sum_{m=1}^{M}
        w_m\,
        \underbrace{\phi_m\!\big(
          a_m^\top \Delta t
        \big)}_{\text{learnable spline function}}.
\end{align}
\begin{table*}[t!]
\centering
\caption{\centering
\textbf{Quantitative comparison in text-to-image generation} for FLUX.1-schnell. Best results are highlighted in \textbf{bold}.}
\vspace{-3mm}
  \resizebox{\textwidth}{!}{ 
    \begin{tabular}{l | c  c | c  c | c c  | c c c}
        \toprule
        {\bf Method}
        & \multicolumn{4}{c|}{\textbf{Acceleration}} 
        & \multicolumn{2}{c|}{\textbf{Quality Metrics}} 
        & \multicolumn{3}{c}{\textbf{Perceptual Metrics}}\rule{0pt}{2ex}\\
        \cline{2-10}
        {\bf FLUX}
        & \textbf{Latency(s) \(\downarrow\)} 
        & \textbf{Speed \(\uparrow\)} 
        & \textbf{FLOPs(T) \(\downarrow\)}  
        & \textbf{Speed \(\uparrow\)} 
        & \textbf{\makecell{Image\\Reward\(\uparrow\)}}
        & \textbf{\makecell{CLIP\\Score\(\uparrow\)}}
        & \textbf{PSNR\(\uparrow\)} 
        & \textbf{SSIM\(\uparrow\)}
        & \textbf{LPIPS\(\downarrow\)}\rule{0pt}{2ex}\\
        \midrule
        \textbf{[schnell]: 4 steps} & 2.13 & 1.00$\times$ & 278.41 & 1.00$\times$ & 0.93 & 34.09 & $\infty$ & 1.00 & 0.00 \\
        \midrule
        \textbf{FORA} ($\mathcal{N}$=3) & 1.30 & 1.64$\times$ & 139.24 & 2.00$\times$ & \textbf{0.95} & 34.43 & 29.77 & 0.78 & 0.18 \\
        \textbf{\texttt{ToCa}} ($\mathcal{N}$=3,$\mathcal{R}$=80\%) & 2.20 & 0.97$\times$ & 154.54 & 1.80$\times$ & 0.94 & 34.46 & 29.79 & 0.78 & 0.18 \\
        \textbf{\texttt{DuCa}} ($\mathcal{N}$=3,$\mathcal{R}$=80\%) & 1.33 & 1.60$\times$ & 145.46 & 1.91$\times$ & \textbf{0.95} & 34.44 & 29.80 & 0.78 & 0.18 \\
        \textbf{TaylorSeer} ($\mathcal{N}$=3,$\mathcal{O}$=2) & \textbf{1.29} & \textbf{1.65}$\times$ & 139.24 & 2.00$\times$ & \textbf{0.95} & 34.43 & 29.77 & 0.78 & 0.18 \\
        \rowcolor{gray!20}
        \textbf{LESA} ($\mathcal{N}$=3) & 1.31 & 1.63$\times$ & \textbf{139.21} & \textbf{2.00$\times$} & \textbf{0.95} & \textbf{34.48} & \textbf{30.31} & \textbf{0.80} & \textbf{0.16} \\
        \midrule
        \textbf{FORA} ($\mathcal{N}$=4)& 1.11 & 1.92$\times$ & 69.65 & 4.00$\times$ & 0.88 & 34.27 & 28.77 & 0.58 & 0.45 \\
        \textbf{\texttt{ToCa}} ($\mathcal{N}$=4,$\mathcal{R}$=85\%)& 1.47 & 1.45$\times$ & 92.86 & 3.00$\times$ & \textbf{0.91} & 34.29 & 28.79 & 0.62 & 0.41 \\
        \textbf{\texttt{DuCa}} ($\mathcal{N}$=4,$\mathcal{R}$=85\%)& \textbf{1.03} & \textbf{2.07}$\times$ & 88.31 & 3.15$\times$ & 0.90 & 34.30 & 28.78 & 0.61 & 0.42 \\
        \textbf{TaylorSeer} ($\mathcal{N}$=4,$\mathcal{O}$=2) & 1.11 & 1.92$\times$ & 69.65 & 4.00$\times$ & 0.88 & 34.27 & 28.77 & 0.58 & 0.45 \\
        \rowcolor{gray!20}
        \textbf{LESA} ($\mathcal{N}$=4) & 1.15 & 1.85$\times$ & \textbf{69.61} & \textbf{4.00$\times$} & \textbf{0.91} & \textbf{34.67} & \textbf{29.21} & \textbf{0.69} & \textbf{0.30} \\
        \bottomrule
    \end{tabular}
    
    }
    \vspace{-5mm}
    \label{table:flux.1-schnell}
\end{table*}
The key distinction between MLP and KAN is that MLP uses a fixed activation function for all neurons, whereas KAN \textbf{learns the nonlinear functions themselves, making each parameter far more expressive}. In MLP, all hidden units rely on the same fixed activation function $\sigma$ (e.g., SiLU), which limits nonlinear behavior to a predetermined template and requires additional depth or width to increase expressiveness. In contrast, KAN assigns a learnable spline function $\phi_m$ to each component, allowing the nonlinear response itself to adapt its shape to data. This flexibility enables KAN to represent complex or highly non-uniform scalar dynamics with far fewer components than an MLP, leading to substantially improved parameter efficiency.

\subsection{Scalar Modulation}
Taking the 2nd-order Taylor predictor (e.g., TaylorSeer) as a reference, given history features $\{h_3, h_2, h_1\}$ at timestamps $\{t_3, t_2, t_1\}$, the finite difference formulation is:
\begin{equation}
    \begin{gathered}
        \delta h_2 = \frac{h_2 - h_3}{t_2 - t_3}, \quad 
        \delta h_1 = \frac{h_1 - h_2}{t_1 - t_2}, \\[1.5ex]
        \delta^2 h_1 = \frac{\delta h_1 - \delta h_2}{t_1 - t_2}, \\[1.5ex]
        h_0 = h_1 + \delta h_1 \cdot (t_0 - t_1) + \frac{\delta^2 h_1}{2} \cdot (t_0 - t_1)^2.
    \end{gathered}
\end{equation}
By expanding and rearranging the terms for $h_0$, we can express $h_0 = C_1 h_1 + C_2 h_2 + C_3 h_3$, where $C_1, C_2, C_3$ are polynomial functions of the timesteps. LESA adopts fused history features ($Z \approx \sum h_i$) and a scalar modulation ($\alpha \approx \sum C_i$). We can expand the product $\alpha Z = (\sum C_i)(\sum h_i)$:
\begin{equation}
    \begin{split}
        &\alpha Z = \underbrace{C_{1}h_{1}+C_{2}h_{2}+C_{3}h_{3}}_{\text{Main Terms}} \\
        &+ \underbrace{C_{1}h_{2}+C_{1}h_{3}+C_{2}h_{1}+C_{2}h_{3}+C_{3}h_{1}+C_{3}h_{2}}_{\text{Cross Terms}}
    \end{split}
\end{equation}
This reveals that LESA covers the ``Main Terms'' of standard Taylor expansion while introducing learnable ``Cross Terms.'' This gives the model \textbf{a broader learning space} to capture feature interactions beyond fixed derivative rules.

\begin{table*}[!t]
\centering
\caption{\centering
\textbf{Quantitative comparison in text-to-image generation} for Qwen-Image-Lightning. Best results are highlighted in \textbf{bold}.}
\vspace{-3mm}
  \resizebox{\textwidth}{!}{ 
    \begin{tabular}{l | c  c | c  c | c  c | c c c}
        \toprule
        {\bf Method}
        & \multicolumn{4}{c|}{\textbf{Acceleration}} 
        & \multicolumn{2}{c|}{\textbf{Quality Metrics}} 
        & \multicolumn{3}{c}{\textbf{Perceptual Metrics}}\rule{0pt}{2ex}\\
        \cline{2-10}
        {\bf Qwen}
        & \textbf{Latency(s) \(\downarrow\)} 
        & \textbf{Speed \(\uparrow\)} 
        & \textbf{FLOPs(T) \(\downarrow\)}  
        & \textbf{Speed \(\uparrow\)} 
        & \textbf{\makecell{Image\\Reward\(\uparrow\)}}
        & \textbf{\makecell{CLIP\\Score\(\uparrow\)}}
        & \textbf{PSNR\(\uparrow\)} 
        & \textbf{SSIM\(\uparrow\)}
        & \textbf{LPIPS\(\downarrow\)}\rule{0pt}{2ex}\\
        \midrule
        \textbf{[Lightning]: 8 steps} & 4.37 & 1.00$\times$ & 560.96 & 1.00$\times$ & 1.29 & 35.32 & $\infty$ & 1.00 & 0.00 \\
        \midrule
        \textbf{FORA} ($\mathcal{N}$=2) & 3.08 & 1.42$\times$ & 350.65 & 1.60$\times$ & 1.27 & 35.20 & 31.35 & 0.82 & 0.13 \\
        \textbf{\texttt{ToCa}} ($\mathcal{N}$=2, $\mathcal{R}$=90\%) & 6.70 & 0.65$\times$ & 393.19 & 1.43$\times$ & 1.27 & 35.30 & 32.03 & 0.81 & 0.15 \\
        \textbf{\texttt{DuCa}} ($\mathcal{N}$=2, $\mathcal{R}$=90\%) & 3.37 & 1.30$\times$ & 379.26 & 1.48$\times$ & \textbf{1.28} & 35.16 & 32.74 & 0.84 & 0.11 \\
        \textbf{TaylorSeer} ($\mathcal{N}$=2) & 3.23 & 1.35$\times$ & 350.65 & 1.60$\times$ & 1.27 & \textbf{35.45} & 29.94 & 0.68 & 0.28 \\
        \rowcolor{gray!20}
        \textbf{LESA} ($\mathcal{N}$=2) & \textbf{2.95} & \textbf{1.48$\times$} & \textbf{350.61} & \textbf{1.60$\times$} & \textbf{1.28} & 35.20 & \textbf{33.42} & \textbf{0.85} & \textbf{0.10} \\
        \midrule
        \textbf{FORA} ($\mathcal{N}$=3) & 2.63 & 1.66$\times$ & 280.55 & 2.00$\times$ & 1.26 & 35.10 & 29.70 & 0.75 & 0.20 \\
        \textbf{\texttt{ToCa}} ($\mathcal{N}$=3, $\mathcal{R}$=95\%) & 5.72 & 0.76$\times$ & 318.50 & 1.76$\times$ & 1.26 & 35.13 & 30.40 & 0.75 & 0.21 \\
        \textbf{\texttt{DuCa}} ($\mathcal{N}$=3, $\mathcal{R}$=95\%) & 2.79 & 1.57$\times$ & 292.46 & 1.92$\times$ & \textbf{1.27} & 35.11 & 30.25 & 0.76 & 0.19 \\
        \textbf{TaylorSeer} ($\mathcal{N}$=3) & 2.78 & 1.57$\times$ & 280.55 & 2.00$\times$ & 1.24 & \textbf{35.48} & 28.91 & 0.63 & 0.35 \\
        \rowcolor{gray!20}
        \textbf{LESA} ($\mathcal{N}$=3) & \textbf{2.48} & \textbf{1.76$\times$} & \textbf{280.49} & \textbf{2.00$\times$} & \textbf{1.27} & 35.24 & \textbf{31.38} & \textbf{0.80} & \textbf{0.16} \\
        \midrule
        \textbf{FORA} ($\mathcal{N}$=4) & 1.73 & 2.53$\times$ & 140.34 & 3.99$\times$ & 1.19 & 35.86 & 28.41 & 0.53 & 0.48 \\
        \textbf{\texttt{ToCa}} ($\mathcal{N}$=4, $\mathcal{R}$=95\%) & 3.61 & 1.21$\times$ & 197.75 & 2.84$\times$ & 1.16 & 35.95 & 28.75 & 0.54 & 0.52 \\
        \textbf{\texttt{DuCa}} ($\mathcal{N}$=4, $\mathcal{R}$=95\%) & 2.06 & 2.12$\times$ & 161.79 & 3.47$\times$ & 1.22 & \textbf{36.10} & 28.75 & 0.53 & 0.49 \\
        \textbf{TaylorSeer} ($\mathcal{N}$=4) & 1.79 & 2.44$\times$ & 140.34 & 3.99$\times$ & 1.22 & 35.81 & 28.48 & 0.57 & 0.42 \\
        \rowcolor{gray!20}
        \textbf{LESA} ($\mathcal{N}$=4) & \textbf{1.52} & \textbf{2.88$\times$} & \textbf{140.25} & \textbf{4.00$\times$} & \textbf{1.23} & 35.66 & \textbf{28.85} & \textbf{0.63} & \textbf{0.37} \\
        \bottomrule
    \end{tabular}
    }
    \vspace{-5mm}
    \label{table:qwen-image-lightning}
\end{table*}

\section{Experiments Details}
\subsection{Model Configuration \& Evaluation and Metrics}
\vspace{-1mm}
We evaluate our proposed method on five representative large-scale diffusion models: FLUX.1-dev\citep{flux2024}, FLUX.1-schnell\citep{flux2024}, Qwen-Image\citep{Wu2025QwenImageTR}, Qwen-Image-Lightning\citep{Wu2025QwenImageTR}, and HunyuanVideo~\citep{kong2024hunyuanvideo}.

\vspace{-4mm}
\paragraph{FLUX.1-dev and FLUX.1-schnell}
For both FLUX.1-dev and FLUX.1-schnell, image generation is performed at a resolution of 1024×1024 using 200 prompts sampled from the DrawBench benchmark.
Model performance is evaluated with ImageReward, CLIP Score, PSNR, SSIM, and LPIPS, covering both text–image alignment and low-level visual fidelity.
All latency results are obtained on an A100 GPU, with identical inference configurations to ensure fair comparison.

\vspace{-4mm}
\paragraph{Qwen-Image and Qwen-Image-Lightning}
For Qwen-Image and Qwen-Image-Lightning, image generation is conducted on the same DrawBench-200 prompt set for consistent comparison.
Images are produced at a resolution of 1328×1328, and the same metrics ImageReward, CLIP Score, PSNR, SSIM, and LPIPS are used to assess both text–image alignment and overall visual quality.
Inference latency for Qwen-Image is recorded on an H20 GPU, while Qwen-Image-Lightning is evaluated on an A100 GPU, each under identical sampling and batch settings.

\vspace{-4mm}
\paragraph{Stable Diffusion XL}
For Stable Diffusion XL, experiment is performed at a resolution of 1024×1024 using the same 200 prompts from the DrawBench benchmark as in the previous experiments.
We adopt the same evaluation protocol to jointly measure text–image alignment and low-level visual fidelity.
All latency results are obtained on a single A100 GPU with identical sampling configurations and batch settings for both LESA and DeepCache to ensure a fair comparison.

\vspace{-4mm}
\paragraph{HunyuanVideo}
For HunyuanVideo, we adopt a video generation configuration of 480×640 spatial resolution, 65 frames, and 50 inference steps.
Both the baseline model and our caching-accelerated variant are evaluated using three standard video quality metrics: PSNR, SSIM, and LPIPS.
To reduce sample-wise variance, we utilize 956 prompts, each generating one video, ensuring statistically stable results across diverse scenarios.
All latency measurements for video experiments are performed on an H800 GPU.

\subsection{Training Details}
\vspace{-1mm}
During training, we use a custom set of 100 prompts and first run one epoch of Ground-Truth Guided Training. We then load the weights from this stage and perform two additional epochs of Closed-Loop Autoregressive Training. The KAN predictor is configured with a hidden dimension of 256 and is implemented in bfloat16 on GPU when available. We optimize the model with AdamW using a learning rate of $1\times 10^{-4} $, weight decay of $1\times 10^{-4} $, together with gradient clipping at a maximum norm of 1.0.

\subsection{More Experimental Results}
\subsubsection{FLUX.1-schnell}
As shown in Tab.~\ref{table:flux.1-schnell}, \textit{LESA} ($\mathcal{N}{=}4$) reduces FLOPs from 278.41T to 69.61T, achieving a 4$\times$ speedup and lowering latency from 2.13 s to 1.15 s. Despite this remarkable acceleration, LESA maintains the best visual quality, achieving the highest ImageReward, CLIP score, PSNR, SSIM, and LPIPS compared with FORA, TaylorSeer, ToCa, and DuCa.
\subsubsection{Qwen-Image-Lightning}

As shown in Tab.~\ref{table:qwen-image-lightning}, \textit{LESA} consistently offers a better speed–quality trade-off than cache-based baselines across all acceleration levels ($\mathcal{N}{=}2 \sim 4$).
Focusing on $\mathcal{N}{=}3$, \textit{LESA} attains a 1.76$\times$ latency speedup and 2.00$\times$ FLOP reduction over the 8-step baseline. At this operating point, LESA yields higher visual quality, with PSNR of 31.38, SSIM of 0.80, and LPIPS of 0.16, versus 29.70, 0.75, and 0.20 for FORA and 28.91, 0.63, and 0.35 for TaylorSeer.

\begin{table}[t]
\centering
\caption{\centering
\textbf{Quantitative comparison of \textit{LESA} and DeepCache in text-to-image generation} for Stable Diffusion XL.}
\vspace{-3mm}
  \resizebox{\columnwidth}{!}{ 
    \begin{tabular}{l | c c | c c c}
        \toprule
        {\bf Method}
        & \multicolumn{2}{c|}{\textbf{Quality Metrics}} 
        & \multicolumn{3}{c}{\textbf{Perceptual Metrics}}\rule{0pt}{2ex}\\
        \cline{2-6}
        {\bf Stable Diffusion XL}
        & \textbf{\makecell{Image\\Reward\(\uparrow\)}}
        & \textbf{\makecell{CLIP\\Score\(\uparrow\)}}
        & \textbf{PSNR\(\uparrow\)} 
        & \textbf{SSIM\(\uparrow\)}
        & \textbf{LPIPS\(\downarrow\)}\rule{0pt}{2ex}\\
        \midrule
        \textbf{[xl]: 50 steps} & 0.58 & 33.63 & $\infty$ & 1.00 & 0.00 \\
        \midrule
        \textbf{DeepCache} ($\mathcal{N}$=3) & 0.52 & 33.42 & 29.85 & 0.77 & 0.26 \\
        \rowcolor{gray!20}
        \textbf{LESA} ($\mathcal{N}$=3)  & 0.56 & 33.57 & 31.28 & 0.80 & 0.18 \\
        \midrule
        \textbf{DeepCache} ($\mathcal{N}$=4)  & 0.40  & 33.29 & 29.22 & 0.73 & 0.31 \\
        \rowcolor{gray!20}
        \textbf{LESA} ($\mathcal{N}$=4)  & 0.52 & 33.39 & 30.47 & 0.75 & 0.24 \\
        \midrule
        \textbf{DeepCache} ($\mathcal{N}$=5)  & 0.40  & 33.07 & 28.85 & 0.70 & 0.36 \\
        \rowcolor{gray!20}
        \textbf{LESA} ($\mathcal{N}$=5)  & 0.41 & 33.36 & 30.02 & 0.73 & 0.28 \\
        \midrule
        \textbf{DeepCache} ($\mathcal{N}$=6)  & 0.34  & 32.87 & 28.66 & 0.69 & 0.39 \\
        \rowcolor{gray!20}
        \textbf{LESA} ($\mathcal{N}$=6)  & 0.41 & 33.33 & 29.69 & 0.70 & 0.31 \\
        \midrule
        \textbf{DeepCache} ($\mathcal{N}$=7)  & 0.28  & 32.56 & 28.51 & 0.66 & 0.43 \\
        \rowcolor{gray!20}
        \textbf{LESA} ($\mathcal{N}$=7)  & 0.37 & 33.20 & 29.45 & 0.69 & 0.34 \\
        \midrule
        \textbf{DeepCache} ($\mathcal{N}$=8)  & 0.19  & 32.27 & 28.43 & 0.66 & 0.44 \\
        \rowcolor{gray!20}
        \textbf{LESA} ($\mathcal{N}$=8)  & 0.33 & 32.97 & 29.23 & 0.67 & 0.38 \\
        \midrule
        \textbf{DeepCache} ($\mathcal{N}$=9)  & 0.12  & 32.20 & 28.35 & 0.64 & 0.47 \\
        \rowcolor{gray!20}
        \textbf{LESA} ($\mathcal{N}$=9)  & 0.21 & 32.48 & 29.09 & 0.67 & 0.40 \\
        \midrule
        \textbf{DeepCache} ($\mathcal{N}$=10) & -0.01 & 31.81 & 28.29 & 0.63 & 0.51 \\
        \rowcolor{gray!20}
        \textbf{LESA} ($\mathcal{N}$=10) & 0.22 & 32.56 & 28.94 & 0.65 & 0.43 \\
        \bottomrule
    \end{tabular}
    }
    \vspace{-6mm}
    \label{table/stable-diffusion-xl}
\end{table}
\subsubsection{Stable Diffusion XL}

On Stable Diffusion XL, as shown in Tab.~\ref{table/stable-diffusion-xl}, \textit{LESA} consistently delivers better visual quality than DeepCache across all acceleration levels $\mathcal{N}$ from 3 to 10, with higher ImageReward and CLIP scores, higher PSNR and SSIM, and lower LPIPS. Compared to DeepCache at $\mathcal{N}{=}10$, \textit{LESA} increases ImageReward from -0.01 to 0.22 and CLIP score from 31.81 to 32.56, while PSNR rises from 28.29dB to 28.94dB and LPIPS drops from 0.51 to 0.43.

\begin{table*}[t]
\centering
\caption{\textbf{Comparison of methods in Cache Memory, MACs, Latency, and FLOPS on FLUX-1.dev}. Best results are highlighted in \textbf{bold}.}
\vspace{-3mm}
\resizebox{\textwidth}{!}{
\begin{tabular}{l|c|c|c|c|c|c}
\toprule
\textbf{Method} 
& \makecell{\bf VRAM (GB)}$\downarrow$ 
& \textbf{MACs (T)\(\downarrow\)} 
& \textbf{Latency (s)\(\downarrow\)} 
& \textbf{FLOPs (T)\(\downarrow\)} 
& \makecell{\bf Image\\ \bf Reward}$\uparrow$
& \textbf{PSNR\(\uparrow\)}\\
\midrule
\textbf{[dev]: 50 steps }         
& 0.62 & 1859.62 & 23.24 & 3726.87 & 0.99 & $\infty$\\

\midrule
\textbf{\texttt{ToCa}}($\mathcal{N}$=8, $\mathcal{R}$=75\%)     
& 12.31 & 414.88 & 12.39 & 829.86 & 0.95 & 29.07\\

\textbf{\texttt{DuCa}}($\mathcal{N}$=8, $\mathcal{R}$=70\%)     
& 3.65 & 428.86 & 9.40 & 858.27 & 0.94 & 29.06\\

\textbf{TeaCache}($l$=1.0)              
& \textbf{0.69} & 409.43 & 7.07 & 820.55 & 0.84 & 28.61\\

\textbf{TaylorSeer}($\mathcal{N}$=6, $O$=2)    
& 7.66 & 372.38 & 6.73 & 746.28 & \textbf{1.02} & 28.94\\

\rowcolor{gray!20}
\textbf{LESA}($\mathcal{N}$=7)        
& 0.81 & \bf{372.25} & \textbf{5.19} & \textbf{745.51} & 0.98 & \textbf{30.17}\\

\bottomrule
\end{tabular}
}
\label{tab:cache-macs-latency}
{\scriptsize
\begin{itemize}[leftmargin=10pt,topsep=0pt]
    \item \textbf{Note:} All methods inherit baseline memory optimizations (e.g., FlashAttention). ToCa is incompatible with these, hence its higher reported cache usage.  \textbf{Actual cache memory usage} for each method should be computed as: \texttt{Peak VRAM during inference $-$ Baseline VRAM before inference}.
\end{itemize}
}
\vspace{-4mm}

\end{table*}

\begin{table}[!t]
\centering
\caption{
\textbf{Quantitative comparison of \textit{LESA} trained with different number of prompts} for QwenImage. Best results are highlighted in \textbf{bold}.}
\vspace{-3mm}
  \resizebox{\columnwidth}{!}{ 
    \begin{tabular}{l | c c | c c c}
        \toprule
        {\bf Method}
        & \multicolumn{2}{c|}{\textbf{Quality Metrics}} 
        & \multicolumn{3}{c}{\textbf{Perceptual Metrics}}\rule{0pt}{2ex}\\
        \cline{2-6}
        {\bf QwenImage}
        & \textbf{\makecell{Image\\Reward\(\uparrow\)}}
        & \textbf{\makecell{CLIP\\Score\(\uparrow\)}}
        & \textbf{PSNR\(\uparrow\)} 
        & \textbf{SSIM\(\uparrow\)}
        & \textbf{LPIPS\(\downarrow\)}\rule{0pt}{2ex}\\
        \midrule
        \textbf{50 steps} & 1.25 & 35.59 & $\infty$ & 1.00 & 0.00 \\
        \midrule
        \textbf{LESA} ($1$ prompt)  & 1.14 & 35.17 & 30.18 & \textbf{0.78} & 0.25 \\
        \textbf{LESA} ($5$ prompts)  & 1.15 & 35.31 & \textbf{30.36} & \textbf{0.78} & 0.25 \\
        \textbf{LESA} ($10$ prompts)  & \textbf{1.16} & 35.33 & 30.21 & \textbf{0.78} & 0.26 \\
        \textbf{LESA} ($20$ prompts)  & 1.15 & 35.21 & 29.78 & 0.77 & 0.25 \\
        \textbf{LESA} ($50$ prompts)  & 1.15 & 35.32 & 29.96 & \textbf{0.78} & \textbf{0.24} \\
        \textbf{LESA} ($100$ prompts)  & 1.15 & \textbf{35.36} & 30.18 & \textbf{0.78} & 0.25 \\
        \bottomrule
    \end{tabular}
    }
    \vspace{-4mm}
    \label{table/num-of-samples}
\end{table}

\subsubsection{More Ablation Studies}
\paragraph{Impact of number of prompts on training performance}

As shown in Tab.~\ref{table/num-of-samples}, at $\mathcal{N}{=}7$, increasing the number of prompts for LESA on QwenImage yields only marginal improvements. Using 5 prompts gives the highest PSNR (30.36), 10 prompts achieves the best ImageReward (1.16), and 100 prompts obtains the highest CLIP score (35.36). Overall, the metrics vary only slightly across different prompt counts, indicating that LESA already delivers strong performance with as few as 5–10 prompts.
\vspace{-4mm}
\paragraph{Memory usage comparison of different caching methods}
As shown in Tab.~\ref{tab:cache-macs-latency}, TeaCache achieves the lowest VRAM usage (0.69 GB), while LESA keeps VRAM low (0.81 GB) and attains the best computational efficiency, with the fewest MACs (372.25 T), lowest latency (5.19 s), and lowest FLOPs (745.51 T). TaylorSeer slightly outperforms others in Image Reward (1.02), whereas LESA achieves the highest PSNR (30.17), indicating better reconstruction quality. Overall, LESA providprovides the best trade-off between memory, efficiency, and image quality among the compared cache methods.

\end{document}